%% file: main.tex
\documentclass[twoside]{article} %

\usepackage[accepted]{aistats2024}

\usepackage[round]{natbib}

\usepackage{hyperref}
\usepackage{url}

\usepackage{mathtools} %

\usepackage{microtype}
\usepackage{graphicx}
\usepackage{subcaption}
\usepackage{booktabs} %

\usepackage{algorithm}
\usepackage{algpseudocode}

\usepackage{amsmath}
\usepackage{amssymb}
\usepackage{mathtools}
\usepackage{amsthm}

\usepackage[capitalize,noabbrev]{cleveref}

\theoremstyle{plain}
\newtheorem{theorem}{Theorem}[section]

\theoremstyle{definition}

\theoremstyle{remark}

\usepackage[utf8]{inputenc} %
\usepackage[T1]{fontenc}    %
\usepackage{hyperref}       %
\usepackage{url}            %
\usepackage{booktabs}       %
\usepackage{amsfonts}       %
\usepackage{nicefrac}       %
\usepackage{microtype}      %
\usepackage{xcolor}         %
\usepackage{siunitx}
\usepackage{varwidth}
\usepackage[font=small,labelfont=bf]{caption}
\usepackage{wrapfig}

\newcommand\citepsupp{\citep}
\newcommand\citetsupp{\citet}

\usepackage{amsthm}

\DeclareMathOperator*{\argmax}{argmax}
\DeclareMathOperator*{\argmin}{argmin}

\newcommand\pistacked{\tilde{\pi}} 
\newcommand\ptruedata{p_{\text{true}}} %

\newcommand\stackedpred{\tilde{\rho}} %
\newcommand\stackedobj{R}
\newcommand\stackedobjest{\hat{\stackedobj}}
\newcommand\stackedobjloo{\stackedobj_{\text{LOO}}}

\newcommand\pacobj{\stackedobj_{\beta}}
\newcommand\latentsdomain{\Theta}
\newcommand\postpred{\rho} %
\newcommand\lppd{\mathrm{LPPD}} %
\newcommand\preddist{p}

\newcommand\gammadist{\Gamma}
\newcommand\invgammadist{\Gamma^{-1}}

\usepackage{listings}
\definecolor{commentgreen}{RGB}{2,112,10}
\definecolor{eminence}{RGB}{108,48,130}
\definecolor{weborange}{RGB}{255,165,0}
\definecolor{frenchplum}{RGB}{129,20,83}
\definecolor{mydarkblue}{rgb}{0,0.08,0.45}
\hypersetup{ %
pdftitle={},
pdfauthor={},
pdfkeywords={},
pdfborder=0 0 0,
pdfpagemode=UseNone,
colorlinks=true,
linkcolor=red,
citecolor=blue,
filecolor=mydarkblue,
urlcolor=mydarkblue,
pdfview=FitH}

\definecolor{todored}{rgb}{0.7,0,0.05}
\definecolor{notegreen}{rgb}{0,0.5,0.1}

\definecolor{stacked}{HTML}{D65244}
\definecolor{stackedval}{HTML}{EE8468}
\definecolor{bma}{HTML}{9EBEFF}
\definecolor{bmaanalytic}{HTML}{7B9FF9}
\definecolor{rjmcmc}{HTML}{5977E3}

\newcommand\colorstacked{\textcolor{stacked}{\textbf{Stacked}}}
\newcommand\colorstackedval{\textcolor{stackedval}{\textbf{Stacked (Val)}}}
\newcommand\colorbma{\textcolor{bma}{\textbf{BMA}}}
\newcommand\colorbmaanalytic{\textcolor{bmaanalytic}{\textbf{BMA (Analytic)}}}
\newcommand\colorrjmcmc{\textcolor{rjmcmc}{\textbf{RJMCMC}}}

\newif\ifdraft
\drafttrue

\ifdraft
\newcommand\todo[1]{\textcolor{todored}{[TODO: #1]}}
\newcommand\note[1]{\textcolor{notegreen}{[NOTE: #1]}}

\else
\newcommand\todo[1]{}
\newcommand\note[1]{}
\newcommand\lo[1]{}
\fi

\begin{document}

\setlength{\abovedisplayskip}{5pt}
\setlength{\belowdisplayskip}{5pt}

\twocolumn[

\aistatstitle{Beyond Bayesian Model Averaging over Paths in Probabilistic Programs with Stochastic Support}

\aistatsauthor{ Tim Reichelt \And Luke Ong \And  Tom Rainforth}
\aistatsaddress{ University of Oxford \And Nanyang Technological University \And University of Oxford} ]

\begin{abstract}
\vspace{-6pt}
The posterior in probabilistic programs with stochastic support decomposes as a weighted sum of the local posterior distributions associated with each possible program path.
We show that making predictions with this full posterior implicitly performs a Bayesian model averaging (BMA) over paths.
This is potentially problematic, as BMA weights can be unstable due to model misspecification or inference approximations, leading to sub-optimal predictions in turn.
To remedy this issue, we propose alternative mechanisms for path weighting: one based on \emph{stacking} and one based on ideas from \emph{PAC-Bayes}.
We show how both can be implemented as a cheap post-processing step on top of existing inference engines. 
In our experiments, we find them to be more robust and lead to better predictions compared to the default BMA weights.
\end{abstract}

\vspace{-6pt}
\section{INTRODUCTION}
\vspace{-4pt}

Universal probabilistic programming systems (PPS) 
\citep{tolpin2016Design,goodman2008Churcha,bingham2019Pyro,ge2018Turing,mansinghka2014Venture}
provide flexible frameworks for expressing powerful probabilistic models, along with tools to aid performing inference in them.
By permitting branching on the outcomes of sampling statements, they allow users to express programs with \emph{stochastic support}, wherein the number of latent variables varies between program executions, leading to challenging inference problems.

Such programs can be thought of as a combination of independent sub-programs, each with static support, known as straight-line programs (SLP)~\citep{chaganty2013Efficiently,sankaranarayanan2013Static,luo2021symbolic}.
The overall posterior is then given by the weighted sum of individual SLP posteriors, with weights corresponding to the local normalization constants of the SLPs; a breakdown recent work has exploited to improve inference~\citep{zhou2020Divide,reichelt2022rethinking}.

We show that this decomposition also reveals that 
the posterior of \emph{any} program with stochastic support is a Bayesian Model Averaging (BMA) \citep{hoeting1999bayesian} over the constituent SLPs of the program.
Thus, all PPS inference engines are implicitly estimating a BMA when the program has stochastic support, whether they explicitly account for this or not.

However, it is widely acknowledged in the Bayesian statistics literature that BMA can be a problematic mechanism for combining the posteriors of individual models \citep{minka2000bayesian,yao2018Using}, with alternatives often preferred in practice, especially when our aim is to make good predictions.
In particular, BMA often performs poorly under \emph{model misspecification}~\citep{gelman2020Holes,oelrich2020bayesian}, wherein it tends to produce
\emph{overconfident} posterior model weights that collapse towards a single model
\citep{huggins2021Reproducible,yang2018Bayesian}.
Given that models will rarely be perfect when working with real data~\citep{box1976science,key1999bayesian,vehtari2012survey}, this is a serious practical concern that has been 
observed to cause notable issues in many applied fields \citep{yang2018Bayesian,smets2007shocks,leff2008cortical}.

We argue that PPSs need to account for these shortfalls and provide access to more robust weighting schemes.
To provide such alternatives, 
we suggest optimizing the SLP weights for \emph{predictive performance}.
Specifically, we introduce weighting schemes based on \emph{stacking of predictive distributions} \citep{wolpert1992stacked,breiman1996stacked,leblanc1996combining,yao2018Using} and \emph{PAC-Bayes objectives} \citep{masegosa2020learning,masiha2021learning,morningstar2022PAC,alquier2023userfriendly}.
We show how to run them as a cheap post-processing step on the outputs of any sample-based inference scheme, and demonstrate that they provide more robust weights with better predictive performance.

Our contributions are:
(a) By interpreting the posterior in programs with stochastic support as a BMA, we show that the weights assigned to SLPs can be unstable, e.g. due to model misspecificiation.
(b) Providing a general scheme to adapt PPS inference algorithms to utilize alternative weighting schemes, with an implementation in Pyro \citep{bingham2019Pyro}. 
(c) Investigating their behaviour for a variety of different programs and showing its benefits on synthetic and real-world data.

\vspace{-5pt}
\section{BACKGROUND}

\vspace{-3pt}
\subsection{Bayesian Model Averaging}
\label{sec:bma_background}
\vspace{-3pt}

An important question in Bayesian statistics is how to best combine the inferences of different possible models.
In a pure Bayesian framework, this is done by weighting the model posteriors according to their \emph{posterior model probability}, leading to a framework called Bayesian model averaging (BMA)~\citep{hoeting1999bayesian}.

To be more precise, assume we have a countable set of Bayesian models indexed by $k$, each with corresponding latent parameters $\theta_k \in \latentsdomain_k$, prior $p_k(\theta_k)$, and likelihood $p_k(y|\theta_k)$, where $y$ is data we want to condition on.\footnote{Note our formulations apply equally when there are also inputs the model is conditioned on, i.e.~we have $p_k(y|\theta_k,x)$, but we negate this from our notation to avoid clutter.}
In BMA, we set a prior over which model generated the data, $p(M=k)$, from which we can derive the posterior model probability
\begin{equation}
    \label{eq:bma_posterior_model_probabilities}
    p(M{=}k \mid y) \propto p(y \mid M{=}k) \, p(M{=}k)
\end{equation}
where $p(y \mid M{=}k) = \int p_k(y|\theta_k) \, p_k(\theta_k) d\theta_k$ is the \emph{model evidence}, or marginal likelihood, for the $k$th model.

Predictions and expectations can now be calculated by combining those from individual models using $p(M{=}k \mid y)$ as weights.
In particular, the posterior predictive distribution for new hypothetical data, $\tilde{y}$, 
is given by
\begin{equation}
    \label{eq:bma_posterior}
    p(\tilde{y} \mid y) \!=\! \sum\nolimits_{k}  p(M{=}k \mid y) \,\mathbb{E}_{p_k(\theta_k|y)} \left[p_k(\tilde{y} | \theta_k)\right],
\end{equation}
where $p_k(\theta_k|y) \propto p_k(\theta_k)p_k(y|\theta_k)$ and $p_k(\tilde{y} | \theta_k)$ are the local posterior and local parameterized predictive distribution, respectively.

\textbf{Criticisms of BMA}~~
In practice, 
our models will never be able to capture the full complexities of the real world as, in the words of George Box, ``all models are wrong, some are useful'' \citep{box1976science}.
It is therefore important to investigate the behaviour of frameworks %
when our models are misspecified, that is when
$\nexists~ (\theta_k, k) : p_k(y | \theta_k) = \ptruedata(y)~ \forall y$, 
where $\ptruedata(y)$ is the (unknown) true data generating distribution.

Crucially, BMA implicitly assumes that the data was sampled from exactly one of the constituent models.
This is often referred to as the $\mathcal{M}$-closed assumption \citep{bernardo2009bayesian,clyde2013bayesian,key1999bayesian}.
As a result, as the amount of data increases the BMA weights will always (except for a few special edge cases) collapse on a single model~\citep{clyde2013bayesian}; the approach reverts to just performing model selection.
Consequently, 
BMA predictions are often inferior compared to other model combination techniques \citep{minka2000bayesian,yao2018Using}.

Viewed another way, model misspecification tends to lead to posterior model probabilities that are \emph{overconfident}:
both empirical and theoretical results have shown that they too readily collapse on a single model~\citep{huggins2021Reproducible,yang2018Bayesian}, even when there are multiple plausible models with similar predictive performance.
Moreover, the exact model onto which the posterior collapses can change drastically when 
regenerating the data from $\ptruedata(y)$.
In general, we expect there to be some variance in the BMA weights due to the fact that we need to \emph{estimate} the model evidence for many real-world models.
However, previous work has demonstrated that overconfidence is an issue even with \emph{analytic} BMA weights \citep{yang2018Bayesian,huggins2021Reproducible,oelrich2020bayesian}. %

\vspace{-5pt}
\subsection{Programs with Stochastic Support}
\label{sec:pps_intro}
\vspace{-3pt}

A probabilistic program can be interpreted as defining an \emph{unnormalized density function} $\gamma : \latentsdomain \rightarrow \mathbb{R}^{\geq 0}$, 
where $\latentsdomain$ denotes the sample space of the latent variables in the program \citep{borgstrom2016lambdacalculus,staton2016Semantics}.
These variables are typically defined as the outcomes of random sampling statements and each sample statement is associated with a unique lexical address.
The goal of inference is then to find a representation of the normalized program density $\pi(\theta) = \gamma(\theta) / \int \gamma(\theta) d\theta$, where $d\theta$ is an implicitly defined reference measure \citep{gordon2014Probabilistic,rainforth2017Automating,vandemeent2018Introduction}. 
One can informally think of $\pi(\theta)$ as a posterior distribution, $p(\theta|y)$.
See App.~\ref{app:ppl_intro} for a more detailed introduction.

Universal PPS allow users to branch on the outcomes of random sampling statements leading to programs with \emph{stochastic support}. %
An important property of such programs is that they can be decomposed into (a countable number of) straight-line programs (SLPs), sub-programs without any control flow \citep{chaganty2013Efficiently,sankaranarayanan2013Static,zhou2020Divide,luo2021symbolic,reichelt2022rethinking}.
These SLPs are effectively the different possible control-flow paths that exist in the program and they are defined by their \emph{address path}, i.e.~the sequence of the lexical addresses encountered during the program's execution.

Each SLP corresponds to a disjoint sub-region, $\latentsdomain_k$, of the overall sample space (such that $\latentsdomain = \bigcup_k \latentsdomain_k$) and
has the \emph{local unnormalized density} $\gamma_k(\theta) := \mathbb{I} \left[ \theta \in \latentsdomain_k \right] \gamma(\theta)$.
The unnormalized density for the whole program can thus be written as $\gamma(\theta) = \sum_{k} \gamma_k(\theta)$.
Similarly, the normalized program density can be rewritten as
\begin{equation}
    \label{eq:ppl_posterior}
    \pi(\theta) = \sum\nolimits_{k} \left(Z_k \big/ \sum\nolimits_{\ell} Z_{\ell} \right) \, \pi_k(\theta), %
\end{equation}
where $Z_k = \int \gamma_k(\theta) d\theta$ and $\pi_k(\theta) = \gamma_k(\theta) / Z_k$
are the \emph{local normalization constant} and \emph{local posterior} respectively.
We refer to $\pi(\theta)$ %
as the \emph{full Bayes posterior}.
Note that the disjoint supports of the SLPs means that there exists exactly one $k : \pi_k(\theta)>0$ for any given $\theta$.

\vspace{-4pt}
\section{INFERENCE IN STOCHASTIC SUPPORT PROGRAMS IS BMA}
\label{sec:inference_as_bma}
\vspace{-4pt}

Examining Eq.~\eqref{eq:ppl_posterior}, we immediately see that $\pi(\theta)$ is a weighted sum of localized posteriors.
This decomposition also reveals that using the full Bayes posterior to calculate predictions or expectations is implicitly performing a BMA over the individual SLPs.
To see this, consider calculating the expectation of some parameterized predictive density $p(\tilde{y}|\theta)$:
\begin{align}
\mathbb{E}_{\pi(\theta)}\left[p(\tilde{y}|\theta)\right]
    = \sum_{k} \frac{Z_k}{ \sum\nolimits_{\ell} Z_{\ell}} \mathbb{E}_{\theta_k\sim\pi_k}\left[p_k(\tilde{y}|\theta_k)\right], \label{eq:ppl_as_bma}
\end{align}
where $p_k$ is any conditional density function such that $p_k(\tilde{y}|\theta)=p(\tilde{y}|\theta) \, \forall \tilde{y}, \theta \in \Theta_k$, and we have defined new random variables $\theta_k$ drawn from the local posterior of the $k$th SLP.
We thus have that the downstream posterior predictive on $\tilde{y}$ is a weighted sum of the posterior predictives that would result from using the $k$th SLP instead of our full program.

There is now a clear analog between Eq.~\eqref{eq:ppl_as_bma} and Eq.~\eqref{eq:bma_posterior}.
To show that the former corresponds to a BMA, all that remains is to show that the weights can be interpreted as posterior model probabilities.
At a high-level, this follows simply from the fact that the $Z_k$ are analogous to (unnormalized) posterior model probabilities (note, though, they are \emph{not} analogous to the model evidences).

To be more precise, consider the factorization $\gamma(\theta)=f(\theta) g(\theta)$ where $f(\theta)$ corresponds to all terms from the sampling and $g(\theta)$ all terms from conditioning statements, such that we can think of them as prior and likelihood components respectively.
The prior probability of choosing the $k$th SLP is now given by $P_k\!:=\!\int f(\theta) \mathbb{I} \left[ \theta \in \latentsdomain_k \right] d\theta$, while the ``model evidence'' is
\begin{align}
    E_k := \int \frac{f(\theta)\mathbb{I} \left[ \theta \in \latentsdomain_k \right]}{P_k} g(\theta) d\theta.
\end{align}
The posterior model probability is then equal to
\begin{align}
    \frac{P_k E_k}{\sum_{\ell} P_{\ell} E_{\ell}} \!=\! \frac{\int f(\theta)\mathbb{I} \left[ \theta \in \latentsdomain_k \right] g(\theta) d\theta}{\sum_{\ell} \int f(\theta)\mathbb{I} \left[ \theta \in \latentsdomain_\ell \right] g(\theta) d\theta} \!=\! \frac{Z_k}{\sum_{\ell} Z_{\ell}}.
\end{align}
Thus Eq.~\eqref{eq:ppl_as_bma} is a BMA with model prior $p(M=k)=P_k$, local posteriors $p_k(\theta_k | y) = \pi_k(\theta_k)$, model evidences $p(y|M=k)=E_k$, and identical local parameterized predictive distributions $p_k(\tilde{y} | \theta_k) = p(\tilde{y}|\theta=\theta_k)$.

Having realized that using the full Bayes posterior leads to BMA, we can instead define a generalized model averaging scheme over the SLPs that is explicitly parameterized by a learnable set of weights, $w$:
\begin{equation}
\label{eq:ppl_posterior_slp_weights}
   \pistacked(\theta; w) := \sum\nolimits_{k} w_k \, \pi_k(\theta)
\end{equation}
with $\sum_{k} w_k = 1, w_k \geq 0$.
This opens the door for considering alternative approaches that avoid the shortfalls of BMA.
We refer to $w_k \propto Z_k$ as the \emph{BMA weights}; the choice made by all current inference engines. 

At this point, a critical reader might argue that all Bayesian inference in general, and not just the weights in a BMA, is sensitive to model misspecification.
However, averaging over a finite discrete set of models has been highlighted as a special case in which Bayesian inference can give counter-intuitive results and is especially susceptible to misspecification \citep{yao2018Using,gelman2020Holes,oelrich2020bayesian}.
Additionally, in most realistic models used in practice we need to estimate the posterior and the local normalization constants.
As we will show in our experiments in Sec.~\ref{sec:experiments} this can be an additional source of variance leading to sub-optimal predictions.
This thus motivates treating the SLP weights %
as explicit parameters that we may wish to set in a non-Bayesian manner, while leaving the local posteriors unchanged.

\vspace{-4pt}
\section{BEYOND BMA PATH WEIGHTS}
\label{sec:stacking_as_alternative}
\vspace{-3pt}

There are different ways we can choose the SLP weights, $w_k$, in Eq.~\eqref{eq:ppl_posterior_slp_weights}, with BMA only one possible choice. 
A simple, albeit crude, alternative would be to just set them equally.
While we find that this can sometimes empirically outperform the BMA weights (see Sec.~\ref{sec:experiments}), it is clearly not an appropriate general-purpose solution and there are cases where it can perform very poorly.

To provide more principled alternatives, we now show how the weights can be optimized to maximize predictive performance. %
For the purpose of exposition, we will introduce the main ideas through the eyes of \emph{stacking} \citep{yao2018Using,wolpert1992stacked,breiman1996stacked,leblanc1996combining} but we will show in Sec.~\ref{sec:pac_bayes} how we can also use PAC-Bayes objectives \citep{morningstar2022PAC} to fit the SLP weights.

\vspace{-3pt}
\subsection{Stacking Objective For PPSs}
\label{sec:stacking}
\vspace{-2pt}

The goal of stacking is to improve predictions by optimizing the model weights, $w$. 
To achieve this, we need to define a method for making predictions for a hypothetical new observation, which we denote as $\tilde{y} \in \mathcal{Y}$. 
Just as we previously defined a generalized version of the posterior in Eq.~\eqref{eq:ppl_posterior_slp_weights}, we now need to establish a generalized version of the posterior predictive.

For simplicity, we will assume for now that an explicit predictive density, $\preddist(\cdot \mid \cdot) : \mathcal{Y} \times \latentsdomain \rightarrow \mathbb{R}^{\geq 0}$, has been provided, before showing how this can instead be derived from the program itself in Sec.~\ref{sec:stacking_as_post}.
This ensures for each SLP we have a \emph{local posterior predictive density}
\begin{equation}
    \postpred_k(\tilde{y}) := \mathbb{E}_{\pi_k(\theta)} \left[ \preddist(\tilde{y} \mid \theta) \right]\label{eq:local_post_pred}. 
\end{equation}
With this, we can define the \emph{stacked predictive density}
\begin{align}
    &\stackedpred(\tilde{y}; w) := \mathbb{E}_{\pistacked(\theta; w)} \left[\preddist(\tilde{y} \mid \theta) \right] = \sum\nolimits_k w_k \, \postpred_k(\tilde{y}) \label{eq:stacked_pred}
\end{align}
In its most general form, stacking defines an objective with a user-defined scoring rule $S(\stackedpred, \tilde{y})$ which takes as input a predictive distribution and a data point~(\citet{gneiting2007strictly}; c.f. App.~\ref{app:scoring_rules}).
It then optimizes the weights, $w$, to maximize the expected score %
\begin{equation}
    \stackedobj(w; S) := \mathbb{E}_{\ptruedata(\tilde{y})} \left[ S(\stackedpred(\cdot \mid w), \tilde{y}) \right].
\end{equation}
We will focus on using the logarithmic score rule, as it is
by far the most popular one used in practice, yielding
\begin{align}
    \label{eq:stacking_objective_log_score}
    \stackedobj(w) :=
    \mathbb{E}_{\ptruedata(\tilde{y})} \left[\log \left( \sum\nolimits_k w_k \, \rho_k(\tilde{y}) \right)\right].
\end{align}
As $\mathbb{E}_{\ptruedata(\tilde{y})} \left[ \log \ptruedata(\tilde{y}) \right]$ is a constant, maximizing $\stackedobj(w)$ is equivalent to minimizing $\mathrm{KL}(\ptruedata(\tilde{y}) \parallel \stackedpred(\tilde{y}; w))$.

We now need a mechanism to estimate the expectation w.r.t.~$\ptruedata(\tilde{y})$ in Eq.~\eqref{eq:stacking_objective_log_score}.
Multiple strategies for this exist \citep{vehtari2012survey}.
We will first show how to do so using an explicit validation set, $\{\tilde{y}_\ell\}_{\ell=1}^{L}$, before describing how to avoid the use of a validation set for a broad class of models in Sec.~\ref{sec:stacking_without_validation}.

As an aside, the stacking weights should be interpreted differently from BMA weights.
In BMA, the weight of the $k$th SLP represents the posterior probability that the data was generated from the $k$th SLP.
In contrast, stacking generates a mixture of the local posterior predictive distributions and optimizes the mixture weights on held-out data.
Hence, the stacking weight for the $k$th SLP estimates the probability that a new data point is drawn from the $k$th SLP.

\vspace{-3pt}
\subsection{Stacking as Post-Processing}
\label{sec:stacking_as_post}
\vspace{-2pt}

Given (normalised) weighted samples $\{(v_s, \theta_s)\}_{s=1}^{S}$ generated from an arbitrary inference algorithm, the posterior of the program is approximated by the empirical measure $\hat{\pi}(\theta) = \sum_{s=1}^{S} v_s \, \delta_{\theta_s}(\theta)$;
unweighted sampling schemes correspond to the special case $v_s = 1 / S$.
The local posteriors of the $k$th SLP are consequently approximated by all the samples which fall into the $k$th SLP, i.e.~$\hat{\pi}_k(\theta) := \sum_{s \in I_k} (v_s / V_k) \, \delta_{\theta_s}(\theta)$ where $I_k := \{s \in \{1, \dots, S\} \mid \theta_s \in \latentsdomain_k\}$ are the indices of the samples from the $k$th SLP and $V_k := \sum_{s \in I_k} v_s$ is the sum of all the associated sample weights.
Recall from Sec.~\ref{sec:pps_intro} that the SLP of a sample $\theta_s$ is determined by its address path.
Thus, we can generate the index sets $I_k$ by grouping all samples with the same path.

\begin{figure}[t]
\vspace{-8pt}
\begin{algorithm}[H]
    \caption{Stacking as Post-Processing (Sec.~\ref{sec:stacking_as_post})}
    \label{alg:stacking}
    \begin{algorithmic}[1]
        \Require Program $\gamma$, Weighted samples from base inference procedure $\{(v_s, \theta_s)\}_{s=1}^{S}$
        \State For each $\theta_s$, record address path and return values $\{g(\tilde{y}_\ell \mid \theta_s)\}_{\ell=1}^{L}$ (c.f. §\ref{sec:stacking_as_post})
        \State Partition sample indices $1, \dots, S$ into subsets $\{I_k\}_{k=1}^K$ using address paths (c.f. §\ref{sec:stacking_as_post})
        \State Compute predictive densities $\hat{\rho}_k(\tilde{y}_{\ell})$ (Eq.~\eqref{eq:pred_density_approx})
        \State Compute $w^* = \argmax \stackedobjest(w)$ (Eq.~\eqref{eq:full_stacking_objective_estimated}) \label{alg_line:objective}
        \State Compute new sample weights $\omega_s$ (Eq.~\eqref{eq:reweighted_samples})
        \State \Return $\{(\omega_s, \theta_s)\}_{s=1}^{S}$
    \end{algorithmic}
\end{algorithm}
\vspace{-25pt}
\end{figure}

The full Bayes posterior hence implicitly uses the approximation $\hat{\pi}(\theta) = \sum_k V_k \, \hat{\pi}_k(\theta)$, assigning the weight $V_k$ to each SLP.
We instead replace these with the learnable SLP weights $w_k$:
\begin{align}
\label{eq:post_stacked}
    \pistacked(\theta; w) &\approx \sum_k w_k \, \hat{\pi}_k(\theta) = \sum_k \sum_{s \in I_k} \frac{w_k \, v_s}{V_k} \, \delta_{\theta_s}(\theta).
\end{align}
We can then use this approximation to get an estimate of the stacked predictive density defined in Eq.~\eqref{eq:stacked_pred}
\begin{align}
    \stackedpred(\tilde{y}_{\ell}; w) \approx& 
    \sum\nolimits_k w_k \, \hat{\rho}_k(\tilde{y}_{\ell}) \\
    \text{where} \quad \hat{\rho}_k(\tilde{y}_{\ell}) :=& \sum\nolimits_{s \in I_k} (v_s/V_k) \, \preddist(\tilde{y}_{\ell} \mid \theta_s)
        \label{eq:pred_density_approx}
\end{align}
are the local posterior approximations.
In our implementation, the user implicitly defines the $\preddist(\tilde{y}_{\ell} \mid \cdot)$ through the program return values.
We can now approximate $\stackedobj(w)$ using
\begin{equation}
    \label{eq:full_stacking_objective_estimated}
    \stackedobjest(w) := \frac{1}{L} \sum\nolimits_{\ell=1}^{L} \log \left( \sum\nolimits_k w_k \, \hat{\rho}_k(\tilde{y}_{\ell}) \right).
\end{equation}
Note that, as the $\hat{\rho}_k(\tilde{y}_{\ell})$ do not depend on the SLP weights,
we can precompute these estimates before optimizing $w$ in a separate, typically cheap, procedure.

After having obtained the optimized weights, $w^* = \argmax_{w} \stackedobjest(w)$, we want to be able to obtain estimates w.r.t. the reweighted normalized density $\pistacked(\theta; w^*)$.
This can be done easily by reweighting the individual posterior samples $\theta_s$.
Letting $k(\theta_s)$ denote the SLP index of sample $\theta_s$ we can rewrite Eq.~\eqref{eq:post_stacked} as
\begin{equation}
    \label{eq:reweighted_samples}
    \pistacked(\theta; w) \approx \sum\nolimits_{s=1}^{S} \omega_s \, \delta_{\theta_s}(\theta);~  \omega_s := \frac{w_{k(\theta_s)} \, v_s}{V_{k(\theta_s)}}.
\end{equation}
Alg.~\ref{alg:stacking} summarizes the high-level steps of our post-processing stacking procedure.
Given an input program and corresponding (weighted) posterior samples, we first use the program to extract the address path and return values for each sample $\theta_s$ 
(e.g.~using the \lstinline{Trace} data structure in Pyro).
Then, we compute the index subsets $I_1, \dots, I_K$ by grouping unique address paths together.
We can then compute the estimate of the local posterior predictive densities $\hat{\rho}_k(\tilde{y}_{\ell})$ and store the estimates in a $K \times L$ matrix.
Finally, this matrix can be used to evaluate our stacking objective $\stackedobjest(w)$ and thus optimize the weights.
This optimization can be done cheaply relative to the cost of inference as 
it is a convex optimization of a small number of parameters and does not require further inferences; we use the L-BFGS-B algorithm \citep{byrd1995limited,zhu1997algorithm} for this.
In App.~\ref{app:implementation} we provide further details on our full implementation in Pyro.

\subsection{Stacking Without Validation Sets}
\label{sec:stacking_without_validation}

So far, we have assumed the existence of an explicit predictive density $\preddist$ and held-out data $\{\tilde{y}_{\ell}\}_{\ell=1}^{L}$, 
but neither is often actually needed to utilize stacking.
Specifically, if we assume that the local unnormalized SLP densities model the observed data $y_i \in \mathcal{Y}$ as conditionally independent given parameters $\theta$, then for each SLP we can write the local unnormalized density as
\begin{equation*}
    \gamma_k(\theta, y_{1:N}) = \mathbb{I}\left[ \theta \in \latentsdomain_k \right]  f(\theta) \prod\nolimits_{i=1}^{N} g(y_i \mid \theta),
\end{equation*}
where %
$f_k : \latentsdomain \rightarrow \mathbb{R}^{\geq 0}$ represents a prior density on $\theta$.
We then use this to derive the following leave-one out (LOO) cross-validation estimator for Eq.~\eqref{eq:stacking_objective_log_score},
\begin{equation*}
    \label{eq:loo_stacking_objectice}
    \stackedobjloo(w) = \, \frac{1}{N} \sum\nolimits_{i=1}^{N} \log \sum\nolimits_k w_k \, \postpred_k(y_i \mid y_{-i}).
\end{equation*}
where $\postpred_k(y_i \mid y_{-i})$ denotes the local posterior predictive density corresponding to $\gamma_k(\theta, y_{1:N}\setminus y_i)$, i.e.~that results from removing the i-th observation term from $\gamma_k$.

Naively computing each of the predictive distributions $\postpred_k(y_i \mid y_{-i})$ would require running inference $N$ times to evaluate the stacking objective.
However, as shown in \cite{yao2018Using}, this can be avoided using Pareto smoothed importance sampling to estimate the LOO densities (PSIS-LOO) \citep{vehtari2017practical}.
Computing the PSIS-LOO approximations to the densities $\postpred_k(y_i \mid y_{-i})$ requires access the individual likelihood terms $g(y_i \mid \theta_s)$ for each posterior sample $\theta_s$ (c.f. App.~\ref{app:psis_explanation} ).
Luckily, these can be extracted automatically for many common PPSs, 
e.g.~using the \lstinline{loo} function of the ArviZ library \citep{kumar2019arviz} for Pyro models.

\vspace{-3pt}
\subsection{Regularized Stacking and PAC-Bayes}
\label{sec:pac_bayes}

Stacking directly optimizes an estimate of the expected predictive density on held-out data (Eq.~\eqref{eq:stacking_objective_log_score}).
Such estimates are fundamentally based on a finite amount of data and %
optimizing them directly
can, at least in principle, lead to overfitting.
Our proposed remedy for this is to add an additional KL regularization term inspired by PAC-Bayes objectives.
Namely we consider
\begin{align*}
    \pacobj(w) &:= \, \frac{1}{L}\sum\nolimits_{\ell=1}^L \log \left(  \sum\nolimits_k w_k \rho_{k}(\tilde{y}_{\ell}) \right) \\
    &- (1 / \beta L) \, \mathrm{KL}(\text{Categorical}(w_1,\dots,w_K) \parallel r(k)),
\end{align*}
where $r(k)$ is a reference weighting we want to regularize towards.
Since we want to discourage the SLP weights from collapsing towards a single SLP, we will generally take $r(k)$ to be the uniform distribution.
The hyperparameter $\beta$ controls the amount of regularization:
$\beta \to \infty$ recovers the standard stacking objective, and $\beta \to 0$ leads to the weights following $r(k)$.

This regularized objective corresponds to a particular instantiation of a PAC-Bayes bound that was proposed by \citet{morningstar2022PAC}.
In the PAC-Bayes literature, $- R(w)$ (Eq.~\eqref{eq:stacking_objective_log_score}) is sometimes referred to as the \emph{true predictive risk}. %
Hence, optimizing $\pacobj(w)$ can be viewed as optimizing a stochastic bound on the true predictive risk, see App.~\ref{app:pac_bayes} for details.

\section{RELATED WORK}

\begin{figure*}[!ht]
  \begin{subfigure}{\textwidth}
    \begin{center}
    \includegraphics[width=0.3\textwidth]{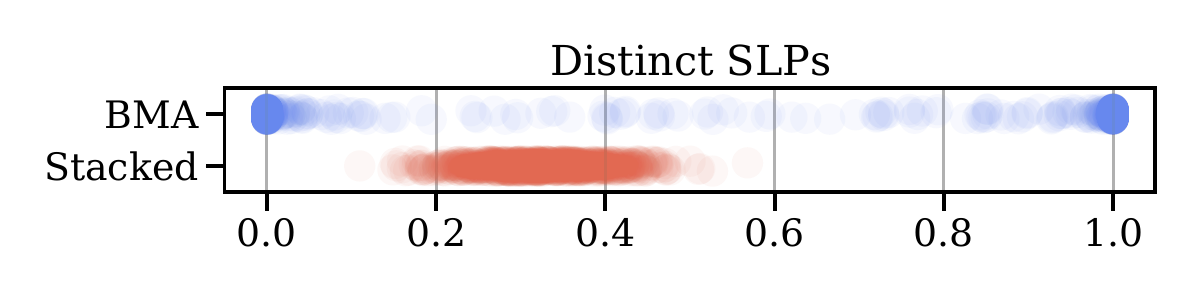}
    ~
    \includegraphics[width=0.3\textwidth]{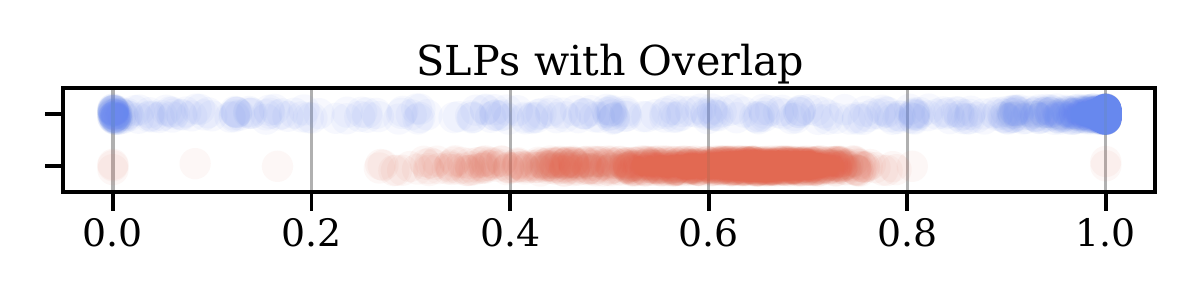}
    ~
    \includegraphics[width=0.3\textwidth]{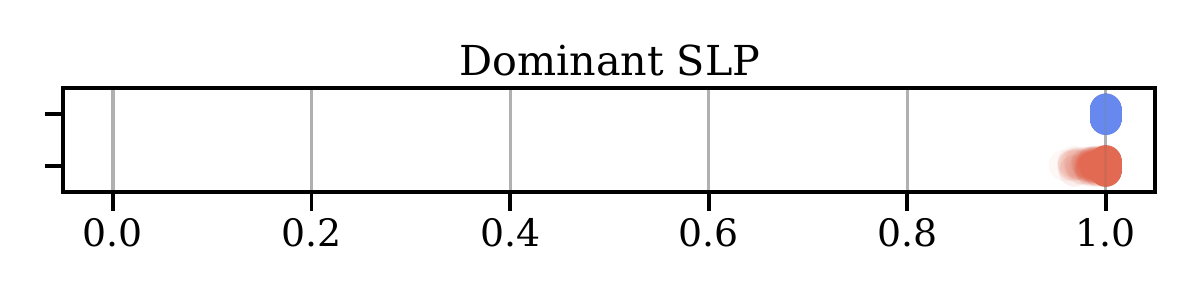}
    \vspace{-7pt}
    \caption{Computed weight for the first SLP; each dot is the weight computed in one of the datasets}
    \label{fig:misspecification_result_weights}
    \end{center}
  \end{subfigure}
  \\
  \begin{subfigure}{\textwidth}
  \begin{center}
  \includegraphics[width=0.3\textwidth, height=40pt, keepaspectratio]{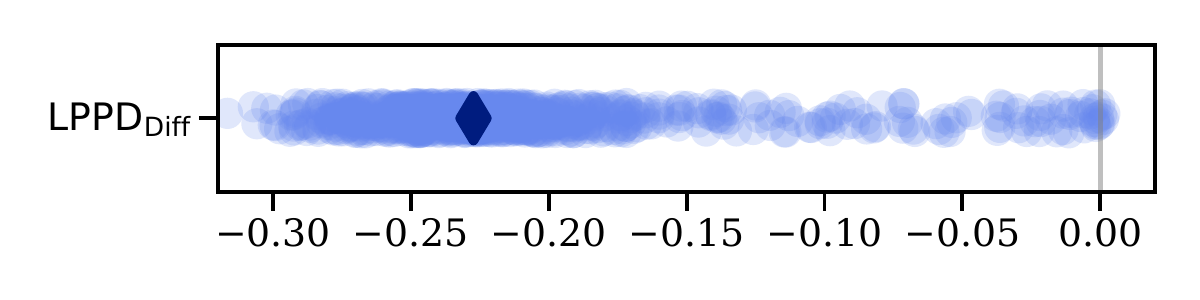}
  ~
  \includegraphics[width=0.3\textwidth, height=40pt, keepaspectratio]{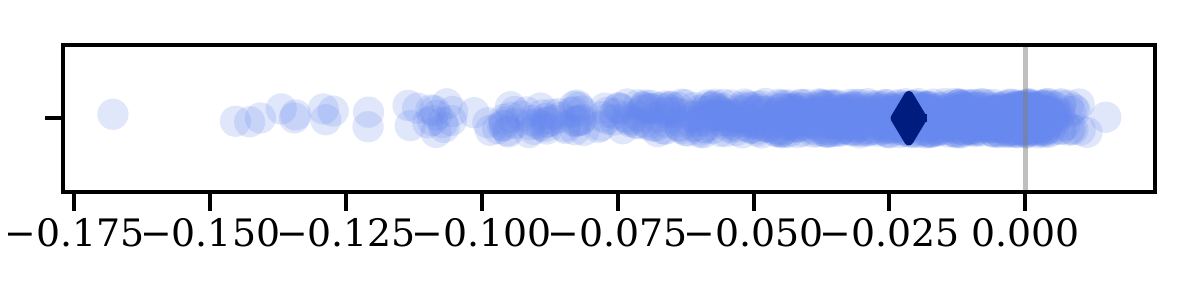}
  ~
  \includegraphics[width=0.3\textwidth, height=40pt, keepaspectratio]{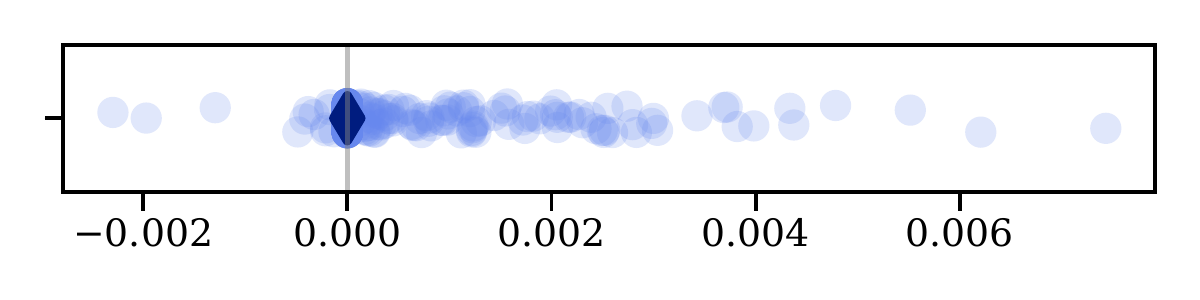}
  \vspace{-7pt}
  \caption{$\lppd_{\text{Diff}}$ on held-out data. Negative values imply stacking is performing better. Diamonds show medians.}
  \label{fig:misspecification_result_lppd}
  \end{center}
  \end{subfigure}
  \vspace{-18pt}
  \caption{
  Behaviour of the BMA and stacked weights in the models as described in Sec.~\ref{sec:misspecification}.
  }
  \label{fig:misspecification_results}
  \vspace{-15pt}
\end{figure*}

\textbf{Alternatives to BMA.}
Bayesian model combination (BMC) \citep{minka2000bayesian,montheith2011turning,kim2012bayesian} aims to break the BMA assumption that the data was generated from exactly one of the candidate models.
This is done by specifying a new extended model which explicitly combines the predictions of all candidate models.
However, fitting the new extended model is significantly more expensive as the inference task no longer breaks down into independent sub-problems. 
Further, how to best combine predictions from different models is highly problem dependent and a modelling decision in its own right and, hence, not suitable for automation. 

\citetsupp{yao2022bayesian} introduced Bayesian hierarchical stacking, which infers different weights for different regions in the covariate space, similar to the frequentist mixture of experts \citepsupp{gormley2019mixture}.
This is less suitable for the fully automated PPS setting because it requires knowledge of the covariate space and assumes the existence of covariates in the first place.
So-called Pseudo-BMA weights \citep{geisser1979predictive} use LOO predictive densities to replace marginal likelihoods but have been shown to work less well than stacking \citep{yao2018Using}.
BayesBag \citep{huggins2021Reproducible} weights models by generating bootstrapped datasets, and averaging the normalization constant over datasets.
This is computationally very intensive as it requires running inference separately in each bootstrapped dataset.

\textbf{PAC-Bayes.}
\citet{warrell2022higher} extend PAC-Bayes bounds to work with hierarchical models inspired by deep probabilistic programs \citep{tran2017deep} and use this extension to derive bounds for multi-task settings such as transfer and meta-learning. 
However, they do not consider programs with stochastic support, the impact of model misspecification, nor integrate their method with a PPS.
PAC-Bayes style arguments have also been used to reason about the hardness of posterior inference in PPS~\citep{freer2010probabilistic}.

\textbf{Programs with stochastic support as BMA.}
Existing inference algorithms for programs with stochastic support (\citet{wingate2011Lightweight,yang2014generating,wood2014New,rainforth2016interacting,le2017Inference,mak2021Nonparametric,mak2022nonparametric}, c.f. App.~\ref{app:ppl_intro}) all implicitly generate a weighting of individual SLPs through the proportion of (weighted) samples generated from each SLP.
Some inference algorithms are adaptions and extensions of the reversible-jump MCMC methods that were originally developed for the BMA setting \citep{green1995Reversible,roberts2019Reversible,cusumano-towner2020Automating}.
However, previous work does not discuss the inherent issues with targeting the BMA model weights and the consequences of this for making predictions; existing algorithms which explicitly assign weights to SLPs only target the default BMA model weights \citep{zhou2020Divide,luo2021symbolic,reichelt2022rethinking}.

\section{EXPERIMENTS}
\label{sec:experiments}

We will now compare different weighting schemes on a range of models and datasets.
Our quantitative measure for comparison will be the average log posterior predictive density (LPPD) on held-out data $\tilde{y}_{1:T}$, where $\mathrm{LPPD} := \frac{1}{T} \sum_{t=1}^{T} \log \stackedpred(\tilde{y}_{t}; w)$.
In particular, we focus on the difference in LPPD from other methods to stacking,
$\lppd_{\text{Diff}} = \lppd_{\text{Other}} - \lppd_{\text{Stacking}}$.

Additionally, we will investigate the behaviour of the SLP weights $w_k$;
a major criticism of the BMA weights is that they are too sensitive to minor changes in the data.
To ensure replicable analysis, we desire the SLP weights to be \emph{robust} and \emph{consistent}, i.e. they should be similar across different possible generated datasets. 

Except when otherwise indicated, we use a variant of the divide, conquer, and combine (DCC) inference algorithm \citep{zhou2020Divide} for the base inference algorithm.
Our DCC implementation uses HMC \citep{neal2011MCMC,hoffman2014no,betancourt2018Conceptual} for the local inference algorithm of each SLP and allocates the computational budget uniformly between SLPs.
Additionally, we also consider the reversible-jump MCMC (RJMCMC, c.f. App.~\ref{app:ppl_intro}) algorithm implemented in Gen \citep{cusumano-towner2019Gen}.
Our implementation is available at \href{https://github.com/treigerm/beyond_bma_in_probprog}{\url{https://github.com/treigerm/beyond_bma_in_probprog}}.

\subsection{When is Stacking Helpful?}
\label{sec:misspecification}
\vspace{-6pt}

\begin{table*}[t]
    \centering
    \caption{
    $\lppd_{\text{Diff}}$ ($\uparrow$ better) for models in Sec.~\ref{sec:subset_reg},~\ref{sec:fun_induction}, and~\ref{sec:var_select}, results computed over 10 replications. Bold indicates no significant difference to Stacked under a Wilcoxon signed-rank test.}
    \label{tab:lppd_results}
    \vspace{-5pt}
    \resizebox{0.97\linewidth}{!}{
      \begin{tabular}{lllllll}
    \toprule %
    \textbf{Model} & \colorstacked & \colorstackedval & \colorbma & \colorbmaanalytic & \colorrjmcmc & \textbf{Equal} \\
    \midrule %
    Subset & $\mathbf{0.0}$ & $-0.01 \pm 0.01$ & $-0.11 \pm 0.05$  & $-0.11 \pm 0.05$ & $-0.11 \pm 0.04$ & $-0.02 \pm 0.01$ \\
    \midrule %
    Fun. Ind. (misspecified) & $\mathbf{0.0}$ & $\mathbf{-1.73\mathrm{e}{-3} \pm 2.98\mathrm{e}{-3}}$ & $\mathbf{-9.10\mathrm{e}{-4} \pm 2.42\mathrm{e}{-3}}$ & N/A & $-0.08 \pm 0.03$ & $-0.31 \pm 0.01$ \\
    Fun. Ind. (well-specified) & $\mathbf{0.0}$ & $\mathbf{-5.61\mathrm{e}{-3} \pm 7.14\mathrm{e}{-3}}$ & $-3.76\mathrm{e}{-1} \pm 2.51\mathrm{e}{-1}$ & N/A & $-2.44 \pm 0.32$ & $-2.31 \pm 0.09$ \\
    \midrule %
    California & $\mathbf{0.0}$ & $-\mathbf{1.55\mathrm{e}{-3} \pm 2.89\mathrm{e}{-3}}$ & $-2.10\mathrm{e}{-2} \pm 6.19\mathrm{e}{-3}$ & $-2.10\mathrm{e}{-2} \pm 6.01\mathrm{e}{-3}$ & $-2.82\mathrm{e}{-1} \pm 1.19\mathrm{e}{-1}$ & $-1.96\mathrm{e}{-1} \pm 3.11\mathrm{e}{-3}$ \\
    Diabetes & $\mathbf{0.0}$ & $-\mathbf{8.22\mathrm{e}{-3} \pm 1.44\mathrm{e}{-2}}$ & $-\mathbf{1.01\mathrm{e}{-2} \pm 1.64\mathrm{e}{-2}}$ & N/A & $-3.83\mathrm{e}{-2} \pm 2.19\mathrm{e}{-2}$ & $-3.66\mathrm{e}{-2} \pm 9.97\mathrm{e}{-3}$ \\
    Stroke & $\mathbf{0.0}$ & $-\mathbf{7.79\mathrm{e}{-4} \pm 1.57\mathrm{e}{-3}}$ & $-6.22\mathrm{e}{-3} \pm 3.81\mathrm{e}{-3}$ & N/A & $-2.25\mathrm{e}{-1} \pm 9.94\mathrm{e}{-2}$ & $-1.31\mathrm{e}{-1} \pm 5.68\mathrm{e}{-3}$ \\
      \bottomrule %
    \end{tabular}
    }
    \vspace{-10pt}
\end{table*}

First, to develop an understanding of the scenarios in which stacking is beneficial, we consider three simple examples in which we limit ourselves to input programs with two SLPs.
Unless otherwise stated, BMA weights are computed analytically and the stacking weights are based on PSIS-LOO.
For each problem, we generate $10^3$ datasets with 200 data points each %
and generate another $10^3$ data points to evaluate the held-out LPPD.

\textbf{Distinct SLPs.} For the first setting, we assume the data is generated by $y_i \sim \mathcal{N}(0, 1)$.
We consider a program with two, misspecified SLPs where the unnormalized density for the $k$th SLP is given by $\gamma_k(\theta_1, \theta_2) = \mathbb{I}\left[\theta_2 = k \right] \frac{1}{2} \prod_{i=1}^{N} \mathcal{N}(y_i; \theta_1, \sigma^2_k) \, \mathcal{N}(\theta_1; 0, 1)$
where we set $\sigma^2_1 = 0.62177$ and $\sigma^2_2 = 2$ (cf. App.~\ref{app:ppl_intro} for the corresponding Pyro program).
This example is adapted from \citet{yang2018Bayesian}.

\textbf{SLPs with overlap.}
Next, we generate a dataset using the relation $y_i = \sum_{d=1}^{4} \beta_d \, x_{i,d} + \epsilon_i$
where $\epsilon_i \sim \mathcal{N}(0, 1)$,  $x_{i,d} \sim \mathcal{N}(0, 1)$ and $\beta = [1.5,1.5,0.3,0.1].$
For inference, we consider a program with two SLPs with unnormalized densities
$\gamma_k(\theta_1, \theta_2, \theta_3) = \mathbb{I}\left[\theta_3 = k \right] \allowbreak \frac{1}{2} \prod_{i=1}^{N} \mathcal{N}(y_i; f_{k}(\theta_1, \theta_2, x_i), 1) \prod_{j=1}^{2} \mathcal{N}(\theta_j; 0, 1)$,
where for the first SLP $f_{1}(\theta_1, \theta_2, x_i) = \theta_1 x_{i,1} + \theta_2 x_{i,3}$ and  for the second $f_{2}(\theta_1, \theta_2, x_i) = \theta_1 x_{i,1} + \theta_2 x_{i,4}$.
Note, that the covariates $x_{i,d}$ are modelled as fixed by the program.
Both SLPs are misspecified because they do not have access to all the covariates. 
However, the two sub-models share complexity/expressiveness, as they both have access to the first covariate.

\textbf{Dominant SLP.} Lastly, we generate data from $y_i = f(x_i)  + \epsilon_i$ with $f(x) = 2 \, x_i + \sin(5 \, x_i)$ and $\epsilon_i, x_{i} \sim \mathcal{N}(0, 1)$.~Our program for inference has two SLPs which are defined as
$\gamma_k(\theta_1, \theta_2, \theta_3) = \mathbb{I}\left[\theta_3 = k \right] \allowbreak\frac{1}{2} \prod_{i=1}^{N} \mathcal{N}(y_i; f_{k}(\theta_1, x_i), \theta_2^2) \, \mathcal{N}(\theta_1; 0, 1) \, \gammadist(\theta_2; 1, 1)$
where $f_{1}(\theta, x) = \theta \, x$, $f_{2}(\theta, x) = \sin(\theta \, x)$, and $\gammadist(\theta; \alpha, \beta)$ is a Gamma distribution parameterized by shape $\alpha$, and rate $\beta$.
The first SLP will provide a significantly better fit to the data as it is able to recover the dominant linear trend.
We use importance sampling to estimate the BMA weights (see App.~\ref{app:experimental_details}).

\textbf{Results.}
The results are presented in Fig.~\ref{fig:misspecification_results}.
For the first two models the stacked weights lead to better predictive performance and more robust weights.
While there is some variation in the stacked weights, their overall distribution is \emph{unimodal} whereas the distribution of the BMA weights are \emph{bimodal} instead.
The tendency of the BMA weights to collapse on either 0 or 1 is exactly the \emph{overconfident} behaviour we described in Sec.~\ref{sec:inference_as_bma}.
In the setting with one dominant SLP, both the BMA and stacking weights lead to similar predictive performance (with BMA doing slightly better) and consistently collapse onto the dominant SLP; here this is desirable behaviour as the first SLP is clearly superior.

\vspace{-5pt}
\subsection{Subset Regression}
\label{sec:subset_reg}
\vspace{-2pt}

\begin{figure}[t]
    \centering
    \includegraphics[width=\textwidth,height=80pt,keepaspectratio]{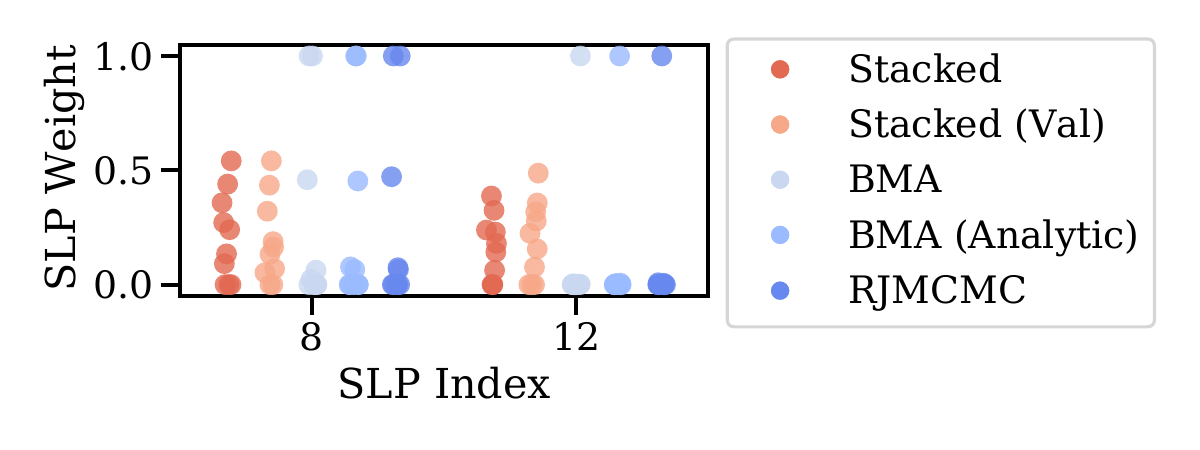}
    \vspace{-15pt}
    \captionof{figure}{
    SLP weights for problem in Sec.~\ref{sec:subset_reg}. 
    Each dot represents the weight of the corresponding SLP in the model. 
    Results are computed over 10 generated datasets. 
    }
    \label{fig:subset_reg_weights}
    \vspace{-19pt}
\end{figure}

To further outline the issues of the default BMA weights, we consider a regression problem with data generated from a linear model of the form $y_i = \epsilon_i + \sum_{d=1}^{15} \beta_d \, x_{i,d}$ where $\epsilon_i \sim \mathcal{N}(0, 1)$ and all the covariates $x_{n,d}$ are drawn independently from $\mathcal{N}(5, 1)$.
Note, the covariates are sampled to generate the synthetic data set, but they are modelled as fixed by the program.
The ground-truth values for the regression coefficients $\beta_d$ are set following a scheme used in \citet{breiman1996stacked} and \citet{yao2018Using} which ensures all covariates are relevant for the prediction of $y_n$ (c.f. App.~\ref{app:experimental_details}).

We compare multiple methods: 1.~\colorstacked, using PSIS-LOO to compute weights, $w$ (Sec.~\ref{sec:stacking_without_validation}); 2.~\colorstackedval, which uses an explicit validation set instead (Sec.~\ref{sec:stacking_as_post}); 2. \colorbma, using the PI-MAIS algorithm \citep{martino2017Layered} to compute local normalization constants (this is the default weighting in DCC); 3.~\colorbmaanalytic, using analytic solutions for the local normalization constants; 4.~\colorrjmcmc, implemented in Gen \citep{cusumano-towner2019Gen};\footnote{We also have conducted experiments that run stacking on top RJMCMC and found that this also led to improvements in predictive performance (c.f. App.~\ref{app:experimental_details}).
Here, we only present the stacking results with samples generated from DCC as this gave the best base inference procedure.} 5. \textbf{Equal}, weights each SLP equally.
We use warm colors to present the results which use one of our proposed alternative objectives to set SLP weights and cooler colours for methods which target the BMA weights.
Note \colorbma, \colorbmaanalytic, and \textcolor{rjmcmc}{\textbf{RJMCMC}} all target the posterior distribution in Eq.~\eqref{eq:ppl_posterior}; in practice, differences between these methods will arise due to differences in the quality of the posterior approximation.

Our input program has 15 SLPs and %
each SLP only gets access to one of the covariates so our overall model is misspecified.
However, since every covariate influences the targets $y_i$, each SLP is relevant for making good predictions.
We generate 50 different datasets from the true data generating distribution, with 200 data points used to run our inferences and $10^3$ data points to evaluate the held-out log posterior predictive density.
For the \textcolor{stackedval}{\textbf{Stacked (Val)}} we use half of the 200 data points for inference and use the rest to estimate the stacking objective (c.f. Eq.~\eqref{eq:full_stacking_objective_estimated}).

Tab.~\ref{tab:lppd_results} shows that stacking outperforms all other methods in terms of predictive performance. 
The BMA weights here actually provide even worse predictions than weighting each SLP equally.
Fig.~\ref{fig:subset_reg_weights} shows the behaviour of the weights for the SLPs with $k= 8$ and $k = 12$ over different randomly generated datasets (see Fig.~\ref{fig:subset_all_weights} in the Appendix for others).
Both \textcolor{bma}{\textbf{BMA}} and \textcolor{bmaanalytic}{\textbf{BMA (Analytic)}} exhibit clear signs of overconfidence as described in Sec.~\ref{sec:bma_background}: the weights often collapse onto a single SLP, but the exact SLP changes between datasets, leading to a bimodal sampling distribution for the SLP weights.
As expected, \textcolor{rjmcmc}{\textbf{RJMCMC}} produces qualitatively and quantitivaly similar results to the other Bayesian weighting mechanisms.
This is in contrast with the \textcolor{stacked}{\textbf{Stacked}} weights which are more evenly spread.
The \textcolor{stackedval}{\textbf{Stacked (Val)}} weights behave qualitatively similarly %
to using PSIS-LOO, but with slightly worse predictive performance.
This is likely due to the corresponding reduction in training set size.

\vspace{-3pt}
\subsection{Function Induction}
\label{sec:fun_induction}
\vspace{-3pt}

\begin{table*}[!ht]
  \begin{minipage}[t]{0.65\linewidth}
    \centering
    \includegraphics[width=\textwidth,height=60pt,keepaspectratio]{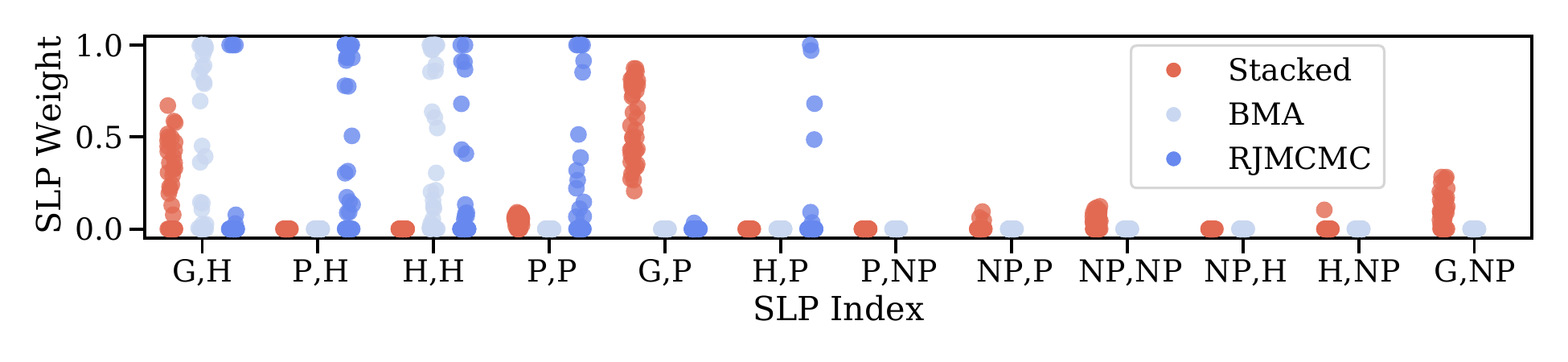}
    \vspace{-10pt}
    \captionof{figure}{
    SLP weights for Sec.~\ref{sec:radon}. 
    X-tick labels indicate the different modelling choices for $\alpha$ and $\beta$; the pattern is ``$\alpha$ model choice, $\beta$ model choice'' with P = pooling, NP = no pooling, H = hierarchical, and G = group-level predictor. 
    }
    \label{fig:radon_weights}
  \end{minipage}
  \hfill
  \begin{minipage}[t]{0.33\linewidth}
    \centering
    \vspace{-55pt}
    \captionof{table}{$\lppd_{\text{Diff}}$ for Radon model.%
    }\label{tab:radon}
    \vspace{-5pt}
    \resizebox{\textwidth}{!}{
    \begin{tabular}{ll}
      \toprule %
      Method & $\lppd_{\text{Diff}}$ ($\uparrow$) \\
      \midrule %
      \colorstacked & $\mathbf{0.0}$ \\
      \colorbma & $-8.24\mathrm{e}{-3} \pm 1.20\mathrm{e}{-2}$ \\
      \colorrjmcmc & $-5.99\mathrm{e}{-2} \pm 1.86\mathrm{e}{-2}$ \\
      \bfseries Equal & $-1.88\mathrm{e}{-2} \pm 1.18\mathrm{e}{-2}$ \\
      \bottomrule %
    \end{tabular}
    }
  \end{minipage}%
  \vspace{-12pt}
\end{table*}

Our next example investigates how well stacking scales to a larger number of SLPs.
We generate observations from the relation $y_i = - x_i + 2 \, \sin(2 \, x_i^2) + \epsilon_i$ with $\epsilon_i \sim \mathcal{N}(0, 0.1^2)$ and the inputs $x_i$ uniformly sampled between -5 and 5. 
We generate 400 data points used for inference and $10^3$ data points for evaluation.
Following \citet{zhou2020Divide}, we use a probabilistic context-free grammar (PCFG) to posit a model over functions.
We consider two PCFGs, 
the \emph{misspecified} PCFG has production rules $e \rightarrow \{ x \, \mid \, \sin(a * e) \, \mid \, a*e + b*e \}$, where $x$ is a terminal symbol denoting an input, and $a$, $b$ are coefficients to be inferred.
The \emph{well-specified} PCFG additionally includes the terminal symbol $x^2$. 
Therefore the well-specified PCFG can express the data generating function whereas the misspecified one cannot.
The program recursively samples from the PCFG using samples from categorical distributions to select production rules and defines latent variables for all the coefficients in a given expression (c.f. App.~\ref{app:experimental_details}).
Note analytic BMA weights cannot be calculated here.

Tab.~\ref{tab:lppd_results} shows that \textcolor{stacked}{\textbf{Stacked}} provides better predictions compared to all other methods, even when the model is well-specified!
Notably, inference in this model is particularly challenging due to the fact that distinct SLPs can have similar or even identical posterior predictive distributions due to symmetries in the PCFG, e.g. $x + \sin(x)$ and $\sin(x) + x$ correspond to two separate SLPs.
The fact that stacking outperforms the methods targeting the BMA weights in the well-specified case is an indicator that the inference algorithms are struggling in this model.
Indeed, we found \textcolor{rjmcmc}{\textbf{RJMCMC}} tends to get stuck in a single SLP which is a well-known issue with MCMC methods for programs with stochastic support (all weights are shown in App.~\ref{app:experimental_details}).
Even though \textcolor{stacked}{\textbf{Stacked}} weights give better predictions, we found that the weights themselves exhibit relatively high-variance.
This is due to the finite sample size of the dataset and the usage of approximate inference algorithms which introduce variance in estimating the stacking objective. %
However, the fact that stacking is able to produce superior predictions shows that it can be a useful mechanism for improving predictive performance even when inference algorithms struggle to produce accurate posteriors approximations.

\vspace{-7pt}
\subsection{Variable Selection}
\label{sec:var_select}
\vspace{-3pt}

Next, we apply stacking to real-world classification and regression tasks.
Here, we have a matrix of covariates $X \in \mathbb{R}^{N \times D}$ and targets $y_{1:N}$, and we want to do variable selection, i.e. select a subset of the features $\mathcal{D} \subseteq \{1, \dots, D\}$ to make predictions.
This problem of variable selection can be encoded as a probabilistic program with stochastic support in which each SLP corresponds to one of the potential subsets of the features $\mathcal{D}$.
We consider three different datasets: \emph{California} (regression) \citep{pace1997sparse}, \emph{Diabetes} (classification) \citep{smith1988using}, and \emph{Stroke} (classification) \citep{kaggle_stroke}.
For the regression, our model is a linear regression with conjugate priors, permitting an analytic solution to the BMA weights.
For the classification tasks, we use a logistic regression model which does not permit an analytic solution.
As the true data generating process in this setting is unknown, we run each method on different train-test splits to estimate the variation in the weights and predictions.

Tab.~\ref{tab:lppd_results} shows the LPPD values of the weighting schemes on different datasets.
The \textcolor{stacked}{\textbf{Stacked}} and \textcolor{stackedval}{\textbf{Stacked (Val)}} weighting schemes generally give better predictive performance compared to the alternatives.
Overall, these results show that stacking can be beneficial for predictive performance even on real-world data.

\vspace{-3pt}
\subsection{Radon Contamination}
\label{sec:radon}
\vspace{-3pt}

Our final example considers the analysis of data about Radon contamination for houses in different US counties \citep{gelman2006data}. 
We here only give a high-level description of the dataset and model, full details in App.~\ref{app:experimental_details}.
For each house recorded in the dataset, we have radon measurements, $y_i$, as well as, a covariate, $x_i \in \{0, 1\}$, which indicates whether the measurement was made in the basement ($x_i = 0$) or first floor of the house ($x_i = 1$).
Our program for this dataset has at its core the regression relation $y_i = \alpha + \beta \, x_i + \epsilon_i$ and the different SLPs in the program make different assumptions for how to model the coefficients $\alpha$ and $\beta$.
For both, we can either: 1) Fit the same coefficient across all counties; 2) Fit a separate coefficient $\alpha_c$ for each county $c$; or 3) Have separate coefficients for each county but assume they come from the same underlying population distribution.
For the intercept term we also consider a fourth option: using county-wide level uranium measurements as a group-level predictor.
The program considers all combinations of modelling choices for $\alpha$ and $\beta$, i.e. it has $4 \cdot 3 = 12$ SLPs.

Each SLP in this program can have potentially hundreds of latent variables and exhibit complex posterior geometries, making this a good testing ground to compare the different weighting schemes.
In Tab.~\ref{tab:radon} we find that the \textcolor{stacked}{\textbf{Stacked}} weights give better performance compared to the \textcolor{bma}{\textbf{BMA}} weights (we do not consider using a validation set here because some counties contain only a handful of observations).
Fig.~\ref{fig:radon_weights} shows that the \textcolor{bma}{\textbf{BMA}} weights tend to concentrate on SLPs with modelling choices ``H,H'' or ``G,H'' 
and have a bimodal sampling distribution.
The \textcolor{stacked}{\textbf{Stacked}} weights are more robust, giving more consistent results between different train-test splits and more conservative weights.
Notably, \textcolor{rjmcmc}{\textbf{RJMCMC}} here collapses onto different SLPs than the BMA weights. 
This is due to the fact that the RJMCMC struggles with SLPs which have a large number of latent variables.
In the limit of infinite samples, the behaviour of RJMCMC and BMA will be identical but Fig.~\ref{fig:radon_weights} illustrates nicely that approximate inference algorithms might collapse onto different SLPs based on the quality of their posterior approximation.

\vspace{-3pt}
\subsection{Impact of Regularization: PAC-Bayes}
\vspace{-3pt}

\begin{figure}[t]
    \centering
    \includegraphics[width=\linewidth,keepaspectratio]{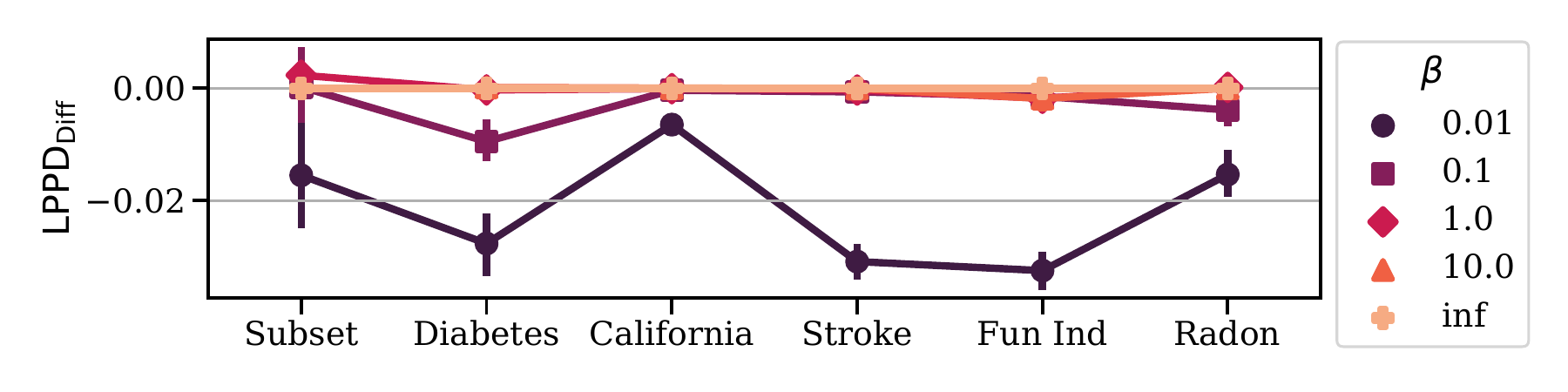}
    \vspace{-22pt}
    \captionof{figure}{
    Impact of regularization parameter $\beta$ on predictive performance in the different models (higher is better).
    Plotted are mean and standard deviation.
    }
    \label{fig:beta_lppd}
    \vspace{-5pt}
\end{figure}

\begin{table}
    \caption{Timings for running inference and stacking, averaged over 5 runs. Inference is conducted using DCC.}
    \label{tab:timings}
    \vspace{-5pt}
    \resizebox{0.97\linewidth}{!}{
      \begin{tabular}{lllll}
    \toprule %
     & Subset & Var. Select. & Fun. Ind. & Radon \\
    \midrule %
    Inference & $29$ s & $297$ s & $285$ m & $700$ s \\
    Stacking & $0.09$ s & $43$ s & $11$ s & $0.2$ s \\
      \bottomrule %
    \end{tabular}
    }
    \vspace{-5pt}
\end{table}

As we have shown in Sec.~\ref{sec:pac_bayes}, the PAC-Bayes bound $\pacobj(w)$ offers an alternative objective to fit the weights and can be interpreted as the stacking loss with an added regularization term where the hyperparameter $\beta$ controls the amount of regularization.
Smaller values of $\beta$ push the stacking weights closer to the uniform distribution over SLPs.
In Fig.~\ref{fig:beta_lppd} we plot the effect of varying $\beta$ on the predictive performance on the different models.
In our experiments, values of $\beta$ below 1 tend to lead to worse predictive performance and the stacking objective with no regularization ($\beta = \infty$ in Fig.~\ref{fig:beta_lppd}) is not outperformed by any form of regularization.
However, depending on the application setting some level of regularization might still be desirable.

\vspace{-5pt}
\section{DISCUSSION} 
\vspace{-3pt}

We have demonstrated that in programs with stochastic support, the conventional posterior path probabilities can be unstable (e.g.~due to model misspecification or inference approximations) and that this in turn can lead to sub-optimal predictions.

In practice, one of the key sources of instability is model misspecification.
When dealing with misspecification, the general advice is to revise or expand the model \citep{gelman2020Bayesian}.
However, when using real-world data it is often not obvious how to further expand a model and mitigate against misspecification.
The radon experiment (Sec.~\ref{sec:radon}) is a good example here as it is already the result of multiple model iterations and expansions, with no clear strategy for how to extend it further.
Additionally, revising and fitting a new expanded model is often prohibitively expensive and therefore not a viable alternative.
As our timings in Tab.~\ref{tab:timings} demonstrate, stacking is a very cheap procedure compared to the cost of inference.
With the automated post-processing techniques presented in this paper, stacking can therefore be conveniently applied at the end of an analysis after the user has gone through multiple iterations of model building.
Thus, rather than viewing stacking as a replacement for model expansion, we view it as a useful tool to safeguard against the instabilities of the default BMA weighting scheme.

While we have demonstrated that the instability in the BMA weights can appear in realistic models and datasets, for any given problem there is no guarantee that the BMA weights will indeed be unstable.
For example, as we saw in the initial experiments in Sec.~\ref{sec:misspecification}, stacking and BMA will produce similar weights if there is one SLP which clearly dominates the others.
However, finding clear criteria that determine when BMA will lead to unstable weights is still an area of open research \citep{yang2018Bayesian,oelrich2020bayesian,huggins2021Reproducible}, so for practitioners it is difficult to know a priori whether a given model will produce unstable SLP weights or not.

Overall, this means there are few reasons not to use stacking: it is cheap, easy-to-use, provides generally more robust weights, and leads to improved predictions.

\subsubsection*{Acknowledgements}

We would like to thank Mrinank Sharma and Andrew Campbell for feedback on earlier drafts of this manuscript. 
We would also like to thank the anonymous reviewers for providing their reviews, especially for encouraging us to explore the connections to PAC-Bayes.
Tim Reichelt is supported by the UK EPSRC CDT in Autonomous Intelligent Machines and Systems with the grant EP/S024050/1. 
Luke Ong acknowledges support from the National Research Foundation, Singapore, under its RSS Scheme (NRF-RSS2022-009).
Tom Rainforth is supported by the UK EPSRC grant EP/Y037200/1.
The authors would like to acknowledge the use of the University of Oxford Advanced Research Computing (ARC) facility in carrying out this work (\url{http://dx.doi.org/10.5281/zenodo.22558}).

\bibliography{references}

\section*{Checklist}

 \begin{enumerate}

 \item For all models and algorithms presented, check if you include:
 \begin{enumerate}
   \item A clear description of the mathematical setting, assumptions, algorithm, and/or model. [Yes]
   \item An analysis of the properties and complexity (time, space, sample size) of any algorithm. [Yes]
   \item (Optional) Anonymized source code, with specification of all dependencies, including external libraries. [Yes]
 \end{enumerate}

 \item For any theoretical claim, check if you include:
 \begin{enumerate}
   \item Statements of the full set of assumptions of all theoretical results. [Not Applicable]
   \item Complete proofs of all theoretical results. [Not Applicable]
   \item Clear explanations of any assumptions. [Not Applicable]     
 \end{enumerate}

 \item For all figures and tables that present empirical results, check if you include:
 \begin{enumerate}
   \item The code, data, and instructions needed to reproduce the main experimental results (either in the supplemental material or as a URL). [Yes]
   \item All the training details (e.g., data splits, hyperparameters, how they were chosen). [Yes]
         \item A clear definition of the specific measure or statistics and error bars (e.g., with respect to the random seed after running experiments multiple times). [Yes]
         \item A description of the computing infrastructure used. (e.g., type of GPUs, internal cluster, or cloud provider). [Yes]
 \end{enumerate}

 \item If you are using existing assets (e.g., code, data, models) or curating/releasing new assets, check if you include:
 \begin{enumerate}
   \item Citations of the creator If your work uses existing assets. [Yes]
   \item The license information of the assets, if applicable. [Yes]
   \item New assets either in the supplemental material or as a URL, if applicable. [Yes]
   \item Information about consent from data providers/curators. [Not Applicable]
   \item Discussion of sensible content if applicable, e.g., personally identifiable information or offensive content. [Not Applicable]
 \end{enumerate}

 \item If you used crowdsourcing or conducted research with human subjects, check if you include:
 \begin{enumerate}
   \item The full text of instructions given to participants and screenshots. [Not Applicable]
   \item Descriptions of potential participant risks, with links to Institutional Review Board (IRB) approvals if applicable. [Not Applicable]
   \item The estimated hourly wage paid to participants and the total amount spent on participant compensation. [Not Applicable]
 \end{enumerate}

 \end{enumerate}

\newpage
\newpage

\input{appendix.tex}

\end{document}

%% file: appendix.tex
\onecolumn

\appendix

\aistatstitle{Beyond Bayesian Model Averaging over Paths in Probabilistic Programs with Stochastic Support}

\section{A Short Introduction to Probabilistic Programming}
\label{app:ppl_intro}

\subsection{Programs as Unnormalized Densities}

\begin{figure}[ht]
    \centering
    \begin{lstlisting}[numbers=none, basicstyle=\footnotesize\ttfamily]
def model(data):
  x = sample("x", Normal(0, 1))
  m1 = sample("m1", Bernoulli(0.5))
  if m1:
    std = 0.62177
    with plate("data1", data.shape[0]):
      sample("y1", Normal(x, std1), obs=data)
  else:
    std = 2.0
    with plate("data2", data.shape[0]):
      sample("y2", Normal(x, std2), obs=data)
    \end{lstlisting}
    \caption{Example Pyro program with stochastic support.} %
    \label{fig:example_program}
\end{figure}

We here give an informal description for how the probabilistic programming language Pyro \citep{bingham2019Pyro} defines unnormalized densities.
For more formal description of the semantic foundations of probabilistic programming we refer the interested reader to \citet{borgstrom2016lambdacalculus,staton2016Semantics} and \citetsupp{lew2019trace}.
Pyro provides the \lstinline{sample} primitive function to both define latent random variables and condition on observed data.
More precisely, calling the function \lstinline{sample(addr, dist)} draws a sample from the distribution object \lstinline{dist} with the corresponding lexical address \lstinline{addr}.
Conditioning on observed data is made possible by passing in the observed data to the \lstinline{sample} function as a keyword argument like so \lstinline{sample(addr, dist, obs=data)}, where again \lstinline{addr} and \lstinline{dist} are assumed to be a lexical address and distribution object.
The users is responsible for ensuring that each \lstinline{sample} statement gets assigned a unique address, i.e. an address that has not been encountered in the current execution, and that every lexical \lstinline{sample} statement within the program has a distinct address.
The second condition avoids the edge case that a program has multiple branches, each of which samples the same sequence of addresses.
This condition is automatically satisfied in some PPS such as Anglican \citep{wood2014New}, however, in PPS in which users manually define addresses they are responsible for adhering to this constraint.
An interesting avenue for future work is automatically verifying that this constraint is satisfied.

Pyro is a universal PPS as it is embedded within the Python programming language and users are free to use language features such as branching on the outcomes of \lstinline{sample} statements.
This means Pyro is able to express programs with stochastic support i.e. the number of \lstinline{sample} statements and their corresponding distribution type can vary from one execution to the next.
Fig.~\ref{fig:example_program} gives an example of a Pyro program with stochastic support.

A Pyro program defines an unnormalized density over the \emph{raw random draws}, $\theta_{1:n_{\theta}} \in \latentsdomain$, which are defined as the sequence of outcomes of the \lstinline{sample} statements (without any observed data) encountered in the program.
Crucially, in models with stochastic support the number of \lstinline{sample} statements $n_x$ can be a random variable and is not fixed for a given program.
Furthermore, the raw random draws are assumed to be the only source of randomness in the program, such that for given raw random draws $\theta'_{1:n_{\theta}}$ all the intermediate variables and return values can be determined deterministically.

Each \lstinline{sample} statement encountered during the program's execution contributes one term to the program's unnormalized density $\gamma(\theta_{1:n_{\theta}})$.
If a \lstinline{sample} does not have associated observed data, then it contributes the term $f_{a_i}(\theta_i \mid \eta_i)$ where $a_i$ is the lexical address, $\theta_i$ the outcome of the \lstinline{sample} statement, $f_{a_i}$ is the parameterized density function of the associated distribution object, and $\eta_i$ are the corresponding parameters.
Similarly, a \lstinline{sample} with observed data $y_j$ contributes the term $g_{b_j}(y_j \mid \phi_j)$ with $b_j$ denoting the lexical address, $g_{b_j}$ the parameterized density function, and $\phi_j$ the parameters.
Overall, as noted by \citep[§4.3.2]{rainforth2017Automating} who formalized this for the PPS Anglican \citepsupp{wood2014New}, this implies the program's unnormalized density function is given by
\begin{equation}
  \label{eq:ppl_density}
  \gamma(\theta_{1:n_{\theta}}) := \prod\nolimits_{i=1}^{n_{\theta}} f_{a_i}(\theta_i \mid \eta_i) \prod\nolimits_{j=1}^{n_y} g_{b_j}(y_j \mid \phi_j).
\end{equation}
\emph{Inference} in a probabilistic program corresponds to the task of finding a representation of the normalized density $\pi(\theta_{1:n}) = \gamma(\theta_{1:n}) / Z$ with normalization constant $Z = \int_{\latentsdomain} \gamma(\theta_{1:n_{\theta}}) d\theta_{1:n_{\theta}}$.

\subsection{Approaches to Inference}

For most problems encountered in practice we cannot exactly compute the normalized program density $\pi(\theta_{1:n_{\theta}})$ and instead have to resort to approximate inference algorithms.
A variety of approaches for approximate inference in probabilistic programs with stochastic support exist.
Particle-based approaches \citepsupp{wood2014New,rainforth2016interacting,murray2020Automated} and algorithms based on importance sampling \citepsupp{le2017Inference,baydin2019Etalumis,harvey2019attention} generate a set of weighted samples as an approximate posterior.
Markov chain Monte Carlo (MCMC) methods with either automatic or manual proposals try to generate a set of samples directly from the posterior \citepsupp{wingate2011Lightweight,yang2014generating,tolpin2015output,roberts2019Reversible,cusumano-towner2020Automating,mak2021Nonparametric,mak2022nonparametric}.
Variational inference algorithms create a parameterized distribution $q(\theta; \phi)$ and optimize the parameters $\phi$ such that $q(\theta; \phi)$ is ``close'' to the full Bayes posterior where closeness is measured with some divergence (most commonly the KL divergence) \citepsupp{wingate2013Automated,ranganath2014Black,paige2016Automatic}.

We now give a more detailed description of involutive MCMC which is the foundation behind Gen's implementation of reversible-jump MCMC (RJMCMC) \citep{cusumano-towner2020Automating}.
In involutive MCMC the user specifies an auxiliary kernel $\kappa$ which acts on an auxiliary variable $v \in Y$, s.t. $\kappa_{\theta}(\cdot) : Y \to [0, \infty)$ for all $\theta \in \latentsdomain$ with $\gamma(\theta) > 0$, and an involution $\Phi : \latentsdomain \times Y \to \latentsdomain \times Y$, i.e. $\Phi^{-1} = \Phi$.
Given an initial state $\theta$, involutive MCMC proceeds by first sampling a new auxiliary variable $v \sim \kappa_{\theta}(\cdot)$, then applying the involution to get the newly proposed state $(\theta, v) \leftarrow \Phi(\theta, v)$, and accepting the new state with the acceptance probability
\begin{equation}
    \alpha := \min \Bigl\{ 1, \frac{\gamma(\theta') \, \kappa_{\theta}(v)}{\gamma(\theta) \, \kappa_{\theta}(v)} \lvert \det (\nabla \Phi(\theta, v)) \rvert \Bigr\}.
\end{equation}
The usage of involutive MCMC is partly automated in Gen. 
The user only has to implement the auxiliary kernel and the involution using a domain-specific language.
Based on these quantities Gen is then able to automatically construct an involutive MCMC kernel \citep{cusumano-towner2020Automating}.

\section{Probabilistic Programs without Predictive Distribution}
\label{app:predictive_dist}

\begin{figure}[ht]
    \centering
    \begin{lstlisting}[numbers=none, basicstyle=\footnotesize\ttfamily]
def model(y):
  # Input data is a list of length 2
  if y[0] > 10:
    x = sample("x1", Normal(10, 1))
    sample("y1", Normal(x, 10), obs=y[0])
    sample("y2", Normal(y[0], sqrt(y[0])), obs=y[1])
  else:
    x = sample("x2", Normal(0, 1))
    sample("y1", Normal(x, 1), obs=y[0])
    sample("y2", Normal(y[0], 1), obs=y[1])
    \end{lstlisting}
    \caption{Example of a probabilistic program without a predictive distribution.}
    \label{fig:program_without_predictive}
\end{figure}

Conventionally, in Bayesian statistics the modeller defines a prior $p(\theta)$ and predictive distribution $p(y_i \mid \theta)$.
Then for a dataset $\mathbf{y} = (y_1, \dots, y_N)$, these two ingredients together define the joint density $p(\theta, \mathbf{y}) = \prod_{i=1}^N p(y_i \mid \theta) \, p(\theta)$ from which can compute the posterior $p(\theta \mid \mathbf{y}) \propto p(\theta, \mathbf{y})$.
In order to predict new data $\tilde{y}$, we can then use the posterior predictive distribution defined as 
\begin{equation}
    p(\tilde{y} \mid \mathbf{y}) = \int p(\tilde{y} \mid \theta) \, p(\theta \mid \mathbf{y}) dx.
\end{equation}
However, more generally, the joint density does not necessarily factorize in this manner which makes it more difficult to automatically deduce a prediction task from the model alone.
Additionally, in the setting of universal probabilistic programming languages the input data can directly influence the model definition as well.
Take for example the Pyro program in Fig.~\ref{fig:program_without_predictive}.
Here, the input data $y$ directly influences what \lstinline{sample} statements we encounter during execution.
Furthermore, in both \lstinline{sample} statenents with address \lstinline{"y2"} the distribution depends on the first data point \lstinline{y[0]}.
Therefore, just from this program definition alone, it is not clear what exactly a reasonable predictive distribution for a new data point would be.

\section{Introduction to Scoring Rules}
\label{app:scoring_rules}

This introduction mainly follows from \citet{gneiting2007strictly} and \citet{yao2018Using}.
\emph{Scoring rules} are functions which take as input a \emph{probabilistic forecast} and a realized event.
The goal of scoring rules is then to evaluate the quality of the probabilistic forecast.
The terminology is specifically used in meteorological forecasts.
In a Bayesian context, these scores are often instead referred to as utilities and Bayesian decision theory aims to maximize the predicted utility of a given action \citep{bernardo2009bayesian}.

More formally, assume we have a random variable on the sample space $(\Omega, \mathcal{A})$ and $\mathcal{P}$ is a convex class of probability measures on $(\Omega, \mathcal{A})$.
Then, any member $P \in \mathcal{P}$ is referred to as a probabilistic forecast and a scoring rule is a function $S : \mathcal{P} \times \Omega \to [-\infty, \infty]$ s.t. $S(P, \cdot)$ is $\mathcal{P}$-quasi-integrable for all forecasts $P \in \mathcal{P}$.
So for a probabilistic forecast $P$ and observed event $y \in \Omega$, $S(P, y)$ is the score which indicates the quality of our forecast.
For notational convenience, if $P$ and $Q$ are both probabilistic forecasts, we define $S(P, Q) = \int S(P, y) dQ(y)$.
Then a scoring rule is \emph{proper} if $S(Q, Q) \geq S(P, Q)$ holds for all $P \in \mathcal{P}$, and \emph{strictly proper} if the equality holds only when $P = Q$ almost surely.

Some common examples of scoring rules include: the \emph{quadratic score} $S(\rho, y) = 2 \rho(y) - \lVert \rho \rVert^2_2$, where $\rho$ is a predictive density; the \emph{logarithmic score} $S(\rho, y) = \log \rho(y)$; and the \emph{continuous-ranked probability score} $S(F, y) = - \int (F(y') - \mathbb{I}[y' \geq y]) dy'$, where $F$ is the cumulative distribution function of the forecast.
Under regularity conditions, \citetsupp{bernardo1979expected} showed that the logarithmic scoring rule is the only proper \emph{local} scoring rule where a local scoring rule is a rule that depends on the predictive density $\rho$ only through the actual observed event $y$.

We refer the reader to \citet{gneiting2007strictly} for more extensive details on scoring rules.

\section{PSIS-LOO Approximation}
\label{app:psis_explanation}

We here only give a brief description of the PSIS-LOO approximation as it is a common procedure.
We refer the reader to \citet{vehtari2017practical} for full details on the approximation and to \citetsupp{vehtari2015pareto} for more details on PSIS.
We want to approximate the local posterior predictive
\begin{align}
    \postpred_k(y_i \mid y_{-i}) &= \int_{\latentsdomain_k} g_k(y_i \mid \theta) \, \pi_k(\theta \mid y_{-i}) d\theta \\
    \intertext{which can be rewritten in terms of the posterior of the full dataset as}
    &= \int_{\latentsdomain_k} g_k(y_i \mid \theta) \, \frac{\pi_k(\theta \mid y_{-i})}{\pi_k(\theta \mid y_{1:N})} \, \pi_k(\theta \mid y_{1:N}) d\theta. \label{eq:loo_integral}
\end{align}
Importantly, the ratio in that integral is proportional to a term that only depends on the individual predictive density
\begin{equation}
    r_i := \frac{1}{g_k(y_i \mid \theta)} \propto \frac{\pi_k(\theta \mid y_{-i})}{\pi_k(\theta \mid y_{1:N})}.
\end{equation}
Hence, we can use a self-normalized importance sampler to get an estimate of Eq.~\eqref{eq:loo_integral} as follows
\begin{equation}
    \postpred_k(y_i \mid y_{-i}) \approx \frac{\sum_{s \in I_k} r^{s}_i \, g_k(y_i \mid \theta_s)}{\sum_{s \in I_k} r^{s}_i}
\end{equation}
where $r^{s}_i = 1 / g_k(y_i \mid \theta_s)$ and $\theta_s$ are (approximate) samples from the posterior. 
However, these ratios $r^{s}_i$ can often have high variance since $\pi_k(\theta \mid y_{1:N})$ will usually have lower variance and thinner tails than $\pi_k(\theta \mid y_{-i})$, in turn, leading to unstable estimates.
To avoid these instabilities, the PSIS-LOO approximation replaces the ratios $r^{s}_i$ with smoothed importance weights $\nu^s_i$.
The importance weights are computed by fitting a generalized Pareto distribution to the raw ratios $r^{s}_i$ and replacing them with the expected order statistics of the fitted Pareto distribution.
This then leads to the estimate
\begin{equation}
    \hat{\postpred}_k(y_i \mid y_{-i}) = \frac{\sum_{s \in I_k} \nu^{s}_i \, g_k(y_i \mid \theta_s)}{\sum_{s \in I_k} \nu^{s}_i}.
\end{equation}

\section{Implementation details}
\label{app:implementation}

\begin{figure}[ht]
    \centering
    \begin{lstlisting}[numbers=none, basicstyle=\footnotesize\ttfamily]
def pyro_subset_regression(X, y, X_val, y_val):
    k = pyro.sample(
        "k", dist.Categorical(torch.ones(X.shape[1]) / X.shape[1]), 
        infer={"branching": True},
    )
    X = X[:, k]
    beta = pyro.sample(f"beta_{k.item()}", dist.Normal(0, np.sqrt(10)))
    sigma = pyro.sample("sigma", dist.Gamma(0.1, 0.1))
    mean = X * beta
    with pyro.plate("data", X.shape[0]):
        pyro.sample("y", dist.Normal(mean, sigma), obs=y)

    X_val = X_val[:, k]
    mean_val = X_val * beta
    return dist.Normal(mean_val, sigma).log_prob(y_val)
    \end{lstlisting}
    \caption{Pyro program for the experiments in Sec.~\ref{sec:subset_reg}.}
    \label{fig:subset_regression_program}
\end{figure}

To be able to evaluate the stacking objective defined in Eq.~\eqref{eq:full_stacking_objective_estimated} we need access to $\hat{\rho}_k(\tilde{y}_{\ell})$, the estimates of the predictive densities.
These, in turn, depend on the evaluations of the predictive densities.
Hence, given validation data $\tilde{y}_1, \dots, \tilde{y}_L$ and posterior samples $\theta_1, \dots, \theta_S$, for stacking we require the evaluations $g(\tilde{y}_{\ell} \mid \theta_s)$.
Note that the local posterior predictive distributions $\rho_k(\tilde{y}_{\ell})$ are expectations under the posterior.

This is important because previous work noted that in the context of probabilistic programming these types of expectations can be formalized as the \emph{expected return values} of a program \citepsupp{gordon2014Probabilistic,zinkov2017Composinga,reichelt2022expectation,lew2023adev}.
Then to apply stacking, we require the user to define a program in which the return values for a given sample $\theta_s$ are the predictive density evaluations $g(\tilde{y}_1 \mid \theta_s), \dots, g(\tilde{y}_L \mid \theta_s)$.
For our Pyro implementation this means that we require the user to define a program which returns an $L$-dimensional vector which correspond to the (log) predictive density on the validation data points.
Fig.~\ref{fig:subset_regression_program} shows how this can be done for the subset regression model from Sec.~\ref{sec:subset_reg}.

To actually compute the stacking weights we rely on Pyro's \lstinline{Trace}\footnote{\url{https://docs.pyro.ai/en/stable/poutine.html\#trace}} data structure which saves important metadata of each program execution such as the distribution type of each \lstinline{sample} statement, the corresponding sampled value, its address, and, crucially, also the return value of the program.
Algorithm~\ref{alg:stacking} can then be implemented as a simple function which takes as input a list of posterior samples in the form of Pyro \lstinline{Trace}s.
Because each \lstinline{Trace} stores the address path of a given run, we can easily determine the set of SLPs and associate each sample with its SLP.
From the \lstinline{Trace} data structure we can then also extract the return values i.e. the evaluations $\log g(\tilde{y}_{\ell} \mid \theta_s)$.
From these evaluations we can calculate the estimates $\hat{\rho}_k(\tilde{y}_{\ell})$ which are needed in our stacking objective in Eq.~\eqref{eq:full_stacking_objective_estimated}.
The optimization of the objective is done using SciPy's \citepsupp{virtanen2020scipy} implementation of the L-BFGS-B optimizer \citepsupp{liu1989limited}.

Similarly, to \citet{zhou2020Divide} and \citet{reichelt2022rethinking} we also allow the user to annotate specific discrete sampling statements to indicate that they influence the SLP of the program.
In our implementation, this is done by passing \lstinline|{"branching": True}| to the \lstinline{infer} keyword argument of Pyro's \lstinline{sample} statement.
Using this annotation the inference backend can then control which SLP of the program is sampled by conditioning these \lstinline{sample} statements on specific values.
Furthermore, these annotations allow the inference backend to deterministically enumerate all SLPs.
In our implementation, we give the user to enumerate all SLPs using breadth-first search, i.e. in the enumeration procedure we run the program as normal but if we encounter an annotated sampling statement we enumerate the whole support of the distribution and put each possible sampled value onto a queue.
Once enumeration of a sample site has finished, execution resumes by popping a value from the front of the queue and continuing executing the program from the \lstinline{sample} statement from which the value was sampled.
This enumeration procedure is similar to how marginalization of discrete sample sites is implemented in Pyro.
Note that we are assuming that each annotated discrete \lstinline{sample} site has finite support.
Future implementations could deal with finite support by truncating the sampling distribution.

\subsection{Alternative Interface for Stacking}

\begin{figure}[ht]
    \centering
    \begin{lstlisting}[numbers=none, basicstyle=\footnotesize\ttfamily]
def model_average(models, model_args, model_kwargs):
    k = pyro.sample(
        "k", dist.Categorical(torch.ones(len(models)) / len(models)), 
        infer={"branching": True},
    )
    return models[k](*model_args, **model_kwargs)
    \end{lstlisting}
    \caption{Alternative interface to enable stacking of a list of user-specified models.}
    \label{fig:alternative_stacking_interface}
\end{figure}

The main aim of our paper is to highlight the shortcomings of the default BMA weights in posteriors of probabilistic programs with stochastic support.
However, as a side-effect our Pyro implementation also provides a convenient mechanism for users to utilize stacking.
Probabilistic programs with stochastic support allow users to encode a large class of problem settings, including the traditional BMA setting.
However, in some cases, for convenience, users might not wish to write a single program with stochastic support which encodes all the different models in the BMA. 
Instead, the user might want to write $K$ separate programs, each with static support, and then want to apply either BMA or stacking to that list of models.
However, this interface is easily encoded within our framework because the user only needs to write another program which combines all the models together.
This program is shown in Fig.~\ref{fig:alternative_stacking_interface}, it takes as input a list of Pyro models, and the arguments that should be passed to these models.
The program then samples an index $k$ and chooses one of the candidate models.
Running standard inference in this program would correspond to BMA.
If the user instead wants to run stacking instead of BMA, they can simply choose our stacking implementation and apply it to this program.

\section{Experimental Details and Additional Results}
\label{app:experimental_details}

To make the stacking approach widely applicable we implemented a version of the DCC framework in Pyro \citep{bingham2019Pyro}.
For our implementation of DCC, we assume that all the stochastic branching happens on variables with discrete support and that the remaining latent variables in the program are continuous.
Models of this form then permit the usage of Pyro's built-in Hamiltonian Monte Carlo \citep{neal2011MCMC,hoffman2014no,betancourt2018Conceptual} as the local inference algorithm.
To improve efficiency, our implementation also leverages Pyro's just-in-time (JIT) compilation ability to compile each SLP.
Note that leveraging Pyro's JIT compilation is possible due to breaking down the original program into its SLPs because by default the JIT compilation cannot handle programs with stochastic support.

Our experiments were conducted on an internal compute cluster which consists of a mix of Intel Broadwell, Haswell, and Cascade Lake CPUs. 
In general, we use 16 cores to parallelize our computation and running a single replication of an experiment finishes in a matter of minutes if not seconds.
An exception is the function induction experiment in Sec.~\ref{sec:fun_induction} which takes around 3 hours for a single replication, using 32 cores. 

\subsection{When is Stacking Helpful?}

For each setting we collect $10^3$ HMC samples from each SLP with $400$ burn-in samples.
To estimate the normalization constant for the dominant SLP setting we use a proposal of $q(\theta_1, \theta_2) = \mathcal{N}(\theta_1; 0, 1) \, \gammadist(\theta_2; 1, 1)$ with $10^6$ samples.

Overall, the experiments show that stacking is particularly useful if there are multiple SLPs which fit the data roughly equally well (with respect to the local normalization constant).
In these cases, the SLP weights are at risk of being unstable and stacking can provide a more robust weight estimation procedure.
On the other hand, if there is one dominant SLP which clearly performs better than all the other SLPs then we would expect there to be less to gain from stacking, as the weights will already be fairly stable.

\subsection{Subset Regression}
\label{app:subset_regression}

\begin{figure}[t]
    \centering
    \includegraphics[width=\textwidth,keepaspectratio]{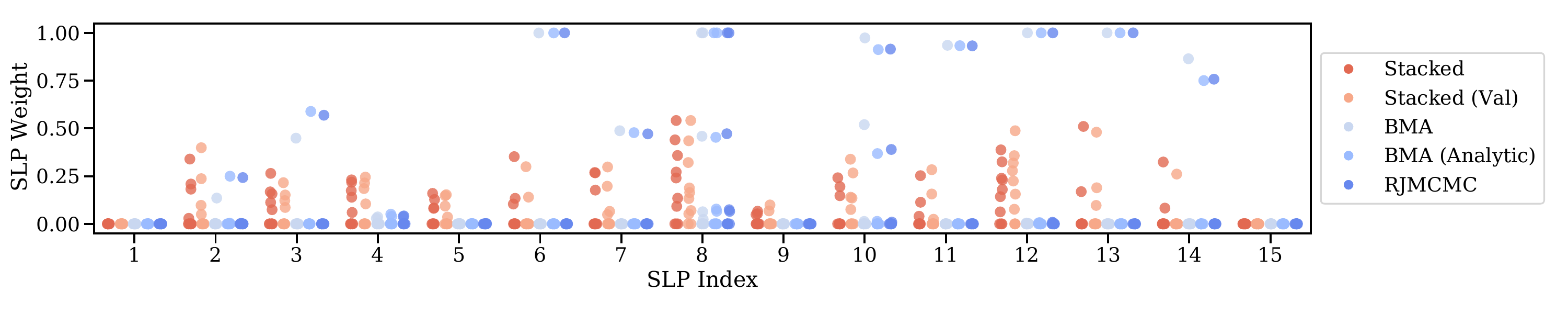}
    \vspace{-20pt}
    \captionof{figure}{
    SLP weights for model in Sec.~\ref{sec:subset_reg}. 
    Each dot represents the weight of the corresponding SLP in the model. 
    Results are computed over 50 randomly generated datasets. 
    }
    \label{fig:subset_all_weights}
\end{figure}

\begin{figure*}[t]
    \centering
    \includegraphics[width=\textwidth,keepaspectratio]{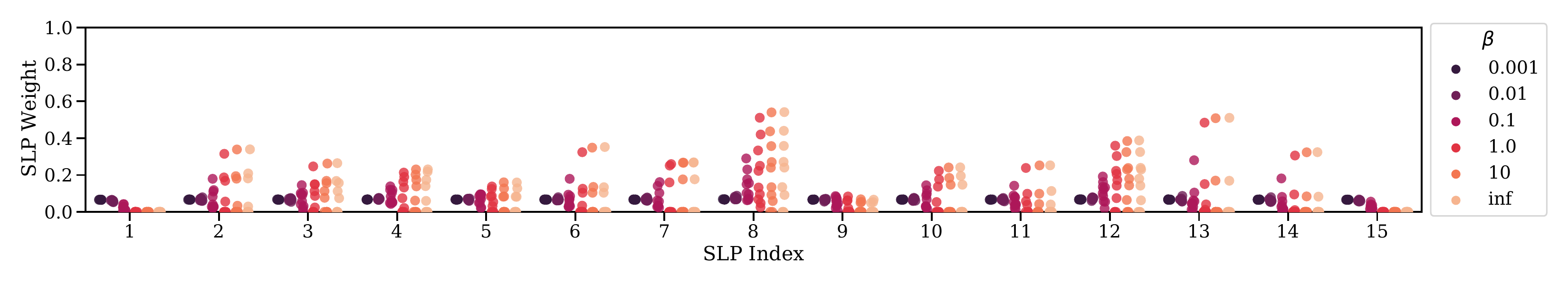}
    \captionof{figure}{Impact of varying regularization parameter $\beta$ on the computed SLP weights. Smaller values of $\beta$ lead to more regularization, pushing the SLP weights more towards the uniform distribution over SLPs.}
    \label{fig:subset_weights_beta}
    \vspace{-10pt}
\end{figure*}

Following \citet{breiman1996stacked} and \citet{yao2018Using}, we generate the regression coefficients according to $\beta_d = \eta \, ( \zeta_d(4) + \zeta_d(8) + \zeta_d(12))$ with $\zeta_d(a) := \mathbb{I}\left[ \lvert d - a < h \rvert \right] \, (h - \lvert d - a \rvert)^2$.
The parameter $h$ determines the number of ``strong'' coefficients.
Following \citet{yao2018Using} we set $h = 5$ which leads to 15 weak coefficients and set $\eta$ such that the signal-to-noise ratio $\mathbb{V}\left[ \sum_{d=1}^{15} \beta_d \, X_{d} \right] / (1 + \mathbb{V}\left[ \sum_{d=1}^{15} \beta_d \, X_{d} \right]) = 4 / 5$ where $X_d$ is the random variable for the $d$th covariate. 
Our input program has 15 SLPs and each SLP has the unnormalized density $\gamma_k(\theta_1, \theta_2, \theta_3)~=~\mathbb{I}\left[\theta_3 = k \right]\allowbreak \, \prod_{i=1}^{N} \mathcal{N}(y_i; \theta_1 x_{i, k}, \theta_2^2) \, \mathcal{N}(\theta_1; 0, 10) \, \gammadist(\theta_2; 0.1, 0.1)\allowbreak \, \text{DiscreteUniform}(\theta_3; 1, 15)$. 
For DCC inference, for each SLP we collect $10^3$ HMC samples with $400$ burn-in samples.
For RJMCMC our transition kernel samples a new $\theta_3$, the variable controlling the covariate that is selected, from a uniform categorical distribution and new $\theta_1$, the local regression coefficient, from a standard normal distribution.
The noise variable is independently updated using a Metropolis-Hastings kernel.
The individual weights for each SLP are shown in Fig.~\ref{fig:subset_all_weights}.

\subsection{Function Induction}

\begin{figure*}[t]
    \centering
    \includegraphics[width=\textwidth,keepaspectratio]{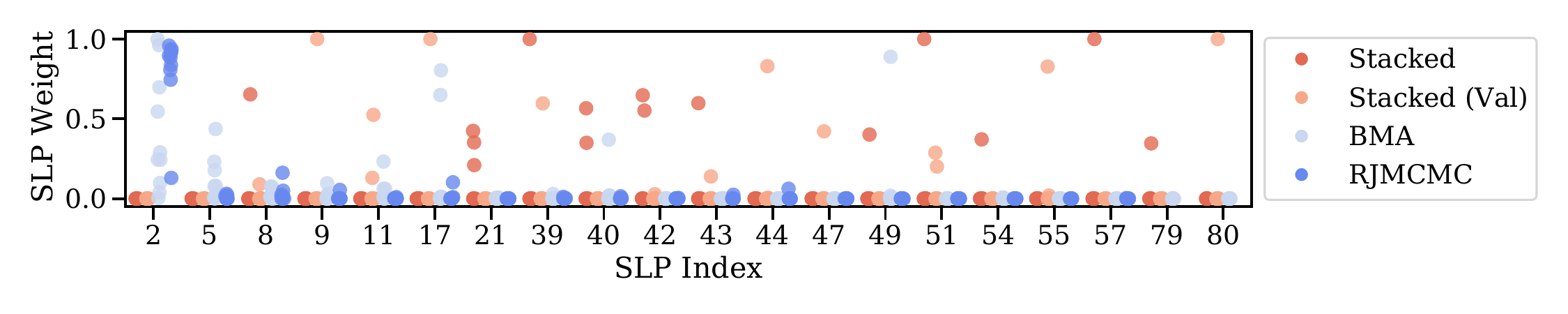}
    \captionof{figure}{SLP weights for the function induction model with the misspecified PCFG. We only plot SLPs which have achieved a weight $> 0.3$ in any run for any method.}
    \label{fig:fun_ind_weights}
    \vspace{-10pt}
\end{figure*}

\begin{figure*}[t]
    \centering
    \includegraphics[width=\textwidth,keepaspectratio]{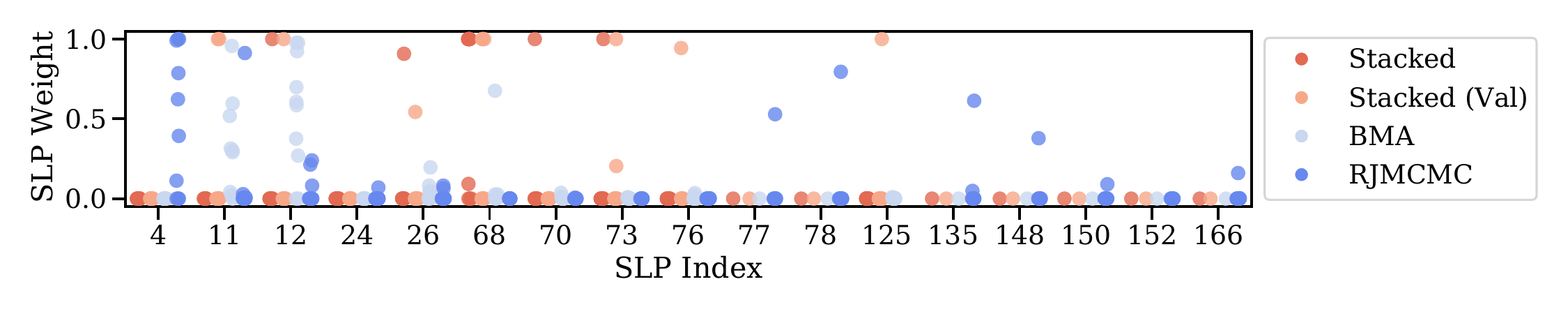}
    \captionof{figure}{SLP weights for the function induction model with the well-specified PCFG. We only plot SLPs which have achieved a weight $> 0.3$ in any run for any method. The SLP indices 11 and 12 both correspond to the true function (due to the symmetry in the $+$ operator there are two SLPs which represent the true function).}
    \label{fig:fun_ind_weights_well_specified}
    \vspace{-10pt}
\end{figure*}

We are using the probabilistic context-free grammar $e \rightarrow \{ x \, \mid \, \sin(a * e) \, \mid \, a*e + b*e \}$.
In our model, we use prior production probabilities $[0.4, 0.4, 0.2]$ and for each of the sampled coefficients (denoted $a$ and $b$ in the PCFG) we use a $\mathcal{N}(0, 10)$ prior.
Using the PCFG we can sample an expression for a function $f : \mathbb{R} \rightarrow \mathbb{R}$, then informally our model can be written as
\begin{align}
    f \sim \text{PCFG}(); \quad  %
    \sigma \sim \Gamma(\sigma; 1, 1); \quad %
    y_i \sim \mathcal{N}(f(x_i), \sigma^2) ~~\text{for}~~ i = 1, \dots, N.
\end{align}
For DCC inference, we run $4$ chains with $500$ HMC samples and $200$ burn-in samples for each SLP.
The sampling distribution of SLP weights are shown in Fig.~\ref{fig:fun_ind_weights}.
For RJMCMC, our transition kernel represents each expression as a PCFG parse tree.
To propose a new expression, the transition kernel picks a random node in the parse tree and replaces that node with a sampled expression from the prior PCFG.
The standard deviation of the likelihood, $\sigma$, is updated independently with a Metropolis-Hastings transition kernel.
For RJMCMC, we collect the same number of total samples as for DCC to ensure a fair comparison.

Due to recursion in the PCFG rules, the program actually defines an infinite number of SLPs.
Similar to \citet{zhou2020Divide}, to avoid infinite recursion we restrict the underlying inference engine to only consider a finite number of SLPs.
Our inference engine enumerates the possible PCFG expressions using breadth-first search and only considers the first 128 expressions.

\subsection{Variable Selection}

\begin{figure}[ht]
    \centering
    \begin{lstlisting}[numbers=none, basicstyle=\footnotesize\ttfamily]
def variable_selection_model(X, y,):
    num_features = X.shape[1]
    features_included = torch.zeros((num_features,), dtype=torch.bool)
    for ix in range(num_features):
        features_included[ix] = pyro.sample(
            f"feature_{ix}", dist.Bernoulli(0.5), infer={"branching": True}
        )
    X_selected = X[:, features_included]
    num_selected = X_selected.shape[1]
    noise_var = pyro.sample("noise_var", dist.InverseGamma(2.0, 1.0))
    with pyro.plate("features", num_selected):
        w = pyro.sample("weights", dist.Normal(0, noise_var.sqrt()))
    means = w @ X_selected.T
    with pyro.plate("data", y.shape[0]):
        pyro.sample("obs", dist.Normal(means, noise_var.sqrt()), obs=y)
    \end{lstlisting}
    \caption{Pyro program for the variable selection experiments.}
    \label{fig:var_select_program}
\end{figure}

\begin{figure*}[!t]
     \centering
     \begin{subfigure}[b]{0.38\textwidth}
         \centering
         \includegraphics[width=\textwidth,height=55pt,keepaspectratio]{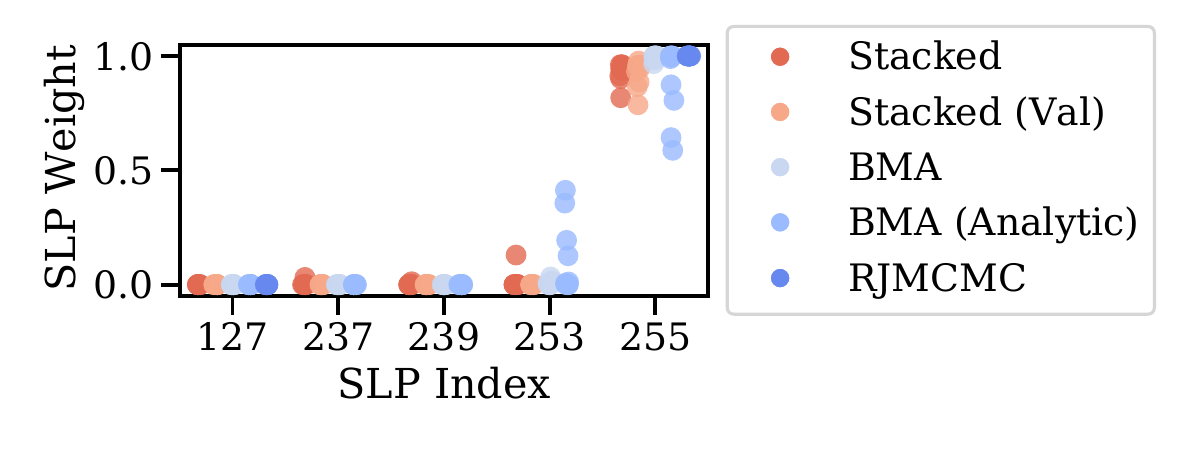}
         \vspace{-10pt}
         \caption{California housing dataset.}
         \label{fig:california_weights}
     \end{subfigure}
     \hfill
     \begin{subfigure}[b]{0.6\textwidth}
         \centering
         \includegraphics[width=\textwidth,height=55pt,keepaspectratio]{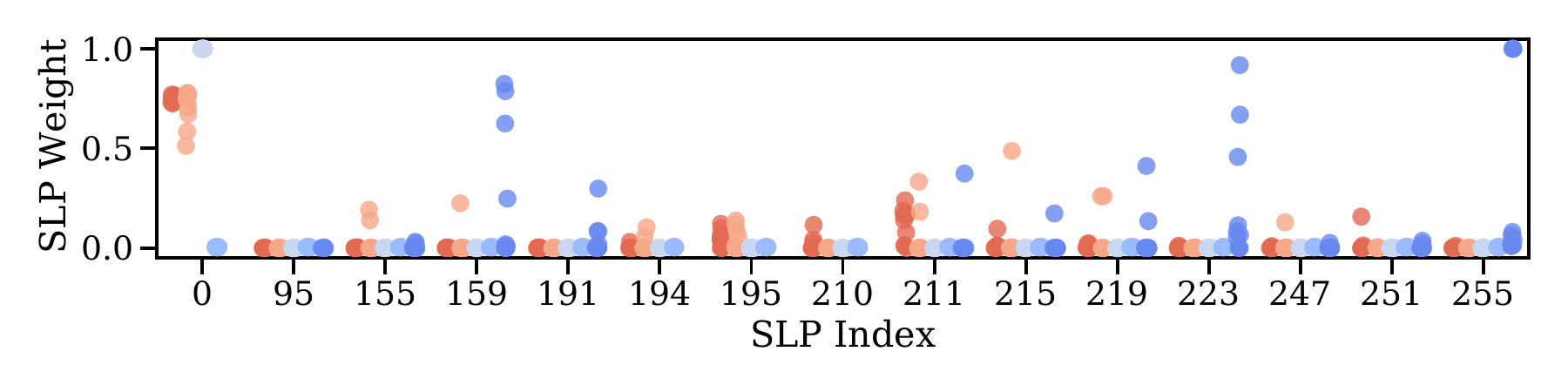}
         \vspace{-10pt}
         \caption{Stroke dataset.}
         \label{fig:stroke_weights}
     \end{subfigure}
     \\
     \begin{subfigure}[b]{\textwidth}
         \centering
         \includegraphics[width=\textwidth,height=55pt,keepaspectratio]{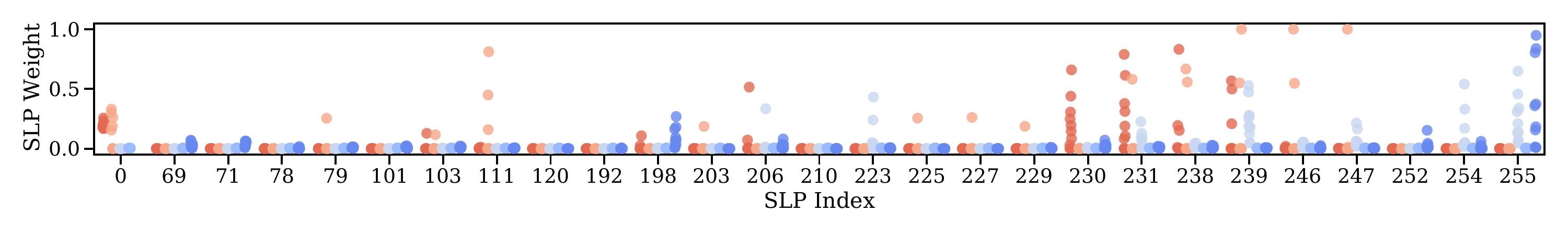}
         \vspace{-10pt}
         \caption{Diabetes dataset.}
         \label{fig:diabetes_weights}
     \end{subfigure}
    \vspace{-15pt}
    \caption{SLP weights for the models in Sec.~\ref{sec:var_select} (conventions as in Fig.~\ref{fig:fun_ind_weights}). The Stroke and Diabetes problems do not permit an analytic solution to the BMA weights.}
    \label{fig:var_select_weights}
    \vspace{-12pt}
\end{figure*}

We assume we are given data $y_{1:N}$ and an associated matrix of covariates $X \in \mathbb{R}^{N \times D}$.
We consider both regression and classification problems. 
The problem of variable selection is to find a subset of the features $\mathcal{D} \subseteq \{1, \dots, D\}$ to make predictions given our data.
For regression task, we have $y_i \in \mathbb{R}$ and our model for a specific subset $\mathcal{D}$ of the features is given by 
\begin{align}
    \sigma^2 &\sim \invgammadist(2, 1), \\
    \beta_d &\sim \mathcal{N}(0, \sigma^2) ~~~~ \text{for } d \in \mathcal{D}, \\
    y_i &\sim \mathcal{N}(\sum\nolimits_{d \in \mathcal{D}} \beta_d x_{i,d}, \sigma^2).
\end{align}
This form of the regression model allows us to analytically calculate the marginal likelihood (see e.g. Ch.~3.5 in \citetsupp{bishop2006pattern}).
For classification tasks, we instead have $y_i \in \{0, 1\}$ and use the logistic regression model
\begin{equation*}
    \beta_d \sim \mathcal{N}(0, 1), \text{ for } d \in \mathcal{D} \text{ and } y_i \sim \mathcal{B}(S(\sum\nolimits_{d \in \mathcal{D}} \beta_d x_{i,d})),
\end{equation*}
where $S(x) = 1 / (1 + \exp(-x))$ and $\mathcal{B}$ is the Bernoulli distribution.

The three different datasets we consider are: \emph{California} housing ($N=20,650$, $D=8$; 50 \% train, 50 \% test) \citep{pace1997sparse}, a regression dataset where the goal is to predict the median house prices for districts of California; \emph{Diabetes} ($N=768$, $D=8$; 80 \% train, 20 \% test) \citep{smith1988using}, a classification dataset on if a person has diabetes or not; and \emph{Stroke} ($N=4908$, $D=8$; 80 \% train, 20 \% test), a classification dataset on if a person will have a stroke or not.
For DCC inference, we collect $10^3$ HMC samples with $400$ burn-in samples for each SLP.
The sampling distribution of the SLP weights are plotted in Fig.~\ref{fig:var_select_weights}.
For RJMCMC, the transition kernel randomly selects a feature dimension $d$ and for that dimension flips the inclusion, i.e. if the feature was previously included it will be excluded and vice versa.
For a previously excluded feature, a new coefficient is sampled from the prior.
For all other coefficients, a new coefficient is proposed from a standard normal centered at the current value.
To ensure a fair comparison to DCC, we collect the same number of samples from RJMCMC as we do for DCC.

\subsection{Modelling Radon Contamination in US Counties}

As mentioned in the main text, our program encodes different modelling choices for both the intercept(s), $\alpha$, and the slope parameter(s), $\beta$, of the regression relation $y_i = \alpha + \beta \, x_i$.
Below we describe in detail the four different modelling choices for the intercept term.
The modelling choices for the slope parameter are analogous, except we do not consider using a group-level predictor for $\beta$.
The full Pyro program for this experiment is shown in Fig.~\ref{fig:radon_program}.

\textbf{Pooling.}
This model corresponds to the SLP denoted ``P, P'' in Fig.~\ref{fig:radon_weights}.
\begin{align}
    \alpha \sim \mathcal{N}(0, 10), ~~
    \beta \sim \mathcal{N}(0, 10), ~~
    f_i = \alpha + \beta \, x_i, ~~
    \sigma \sim \text{Exponential}(5), ~~
    y_i \sim \mathcal{N}(f_i, \sigma^2).
\end{align}

\textbf{No pooling.}
Here $c[i]$ refers to the county index of the $i$th house.
This model corresponds to the SLP denoted ``NP, P'' in Fig.~\ref{fig:radon_weights}.
\begin{align}
    \alpha_c \sim \mathcal{N}(0, 10)~ \text{for each county } c, ~~
    \beta \sim \mathcal{N}(0, 10), ~~
    \sigma \sim \text{Exponential}(5), ~~
    y_i \sim \mathcal{N}(f_i, \sigma^2)
\end{align}
with $f_i = \alpha_{c[i]} + \beta \, x_i$.

\textbf{Hierarchical.}
This model corresponds to the SLP denoted ``H, P'' in Fig.~\ref{fig:radon_weights}.
We are using a non-centered parameterization to allow for better sampling performance from the HMC sampler.
\begin{align}
    \sigma_{\alpha} &\sim \text{Exponential}(1), &
    \mu_{\alpha} &\sim \mathcal{N}(0, 10), &
    \epsilon_c &\sim \mathcal{N}(0, 1)~~~ \text{for each county } c, &
    \alpha_c &= \mu_{\alpha} + \sigma_{\alpha} \, \epsilon_{c}, \\
    \beta &\sim \mathcal{N}(0, 10), &
    \sigma &\sim \text{Exponential}(5), &
    f_i &= \alpha_{c[i]} + \beta \, x_i, &
    y_i &\sim \mathcal{N}(f_i, \sigma^2)
\end{align}

\textbf{Group-level predictor.}
This model corresponds to the SLP denoted ``G, P'' in Fig.~\ref{fig:radon_weights}.
\begin{align}
    \gamma_0 &\sim \mathcal{N}(0, 10), &
    \gamma_1 &\sim \mathcal{N}(0, 10), &
    \sigma_{\alpha} &\sim \text{Exponential}(1), &
    \epsilon_c &\sim \mathcal{N}(0, 1)~~~ \text{for each county } c, \\
    \alpha_c &= \mu_{\alpha,c} + \sigma_{\alpha} \, \epsilon_{c}, &
    \beta &\sim \mathcal{N}(0, 10), &
    \sigma &\sim \text{Exponential}(5), &
    y_i &\sim \mathcal{N}(f_i, \sigma^2)
\end{align}
where $\mu_{\alpha,c} = \gamma_0 + \gamma_1 \, u_c$ and $f_i = \alpha_{c[i]} + \beta \, x_i$.

To ensure we have a balanced representation of data in each county in both the training and the testing data we apply stratified sampling: for each county, we hold out 20 \% of the observations for evaluation.
For this dataset we do not run stacking with a validation set because of the limited amount of data available per county.
For DCC inference, we collect $2000$ HMC samples with $2000$ samples for burn-in for each SLP.
For RJMCMC, the transition kernel picks a modelling choice of $\alpha$ and $\beta$ and then samples the local parameters for each modelling choice from the prior.
The standard deviation $\sigma$ is updated separately with a Metropolis-Hastings transition kernel.
To ensure a fair comparison between DCC and RJMCMC we collect the same total number of samples.

\begin{figure}[ht]
    \centering
    \begin{lstlisting}[numbers=none, basicstyle=\footnotesize\ttfamily]
def radon_model(log_radon, floor_ind, county, num_counties, uranium):
    alpha_choice = pyro.sample(
        "alpha_choices", dist.Categorical(torch.ones(4) / 4), infer={"branching": True}
    )
    if alpha_choice == 0:
        # Pooled model
        alpha = pyro.sample("alpha", dist.Normal(0, 10))
    elif alpha_choice == 1:
        # County specific intercepts
        with pyro.plate("num_alpha", num_counties):
            alpha = pyro.sample("alpha", dist.Normal(0, 10))
        alpha = alpha[..., county]  # Shape: (num_counties,) -> (num_data,)
    elif alpha_choice == 2 or alpha_choice == 3:
        if alpha_choice == 2:
            # Partially pooled model
            mean_a = pyro.sample("mean_a", dist.Normal(0, 1))
        elif alpha_choice == 3:
            # Uranium context
            gamma_0 = pyro.sample("gamma_0", dist.Normal(0, 10))
            gamma_1 = pyro.sample("gamma_1", dist.Normal(0, 10))
            mean_a = gamma_0 + gamma_1 * uranium

        std_a = pyro.sample("std_a", dist.Exponential(1))
        with pyro.plate("num_alpha", num_counties):
            z_a = pyro.sample("z_a", dist.Normal(0, 1))
        alpha = mean_a + std_a * z_a
        alpha = alpha[..., county]  # Shape: (num_counties,) -> (num_data,)

    beta_choice = pyro.sample(
        "beta_choices", dist.Categorical(torch.ones(3) / 3), infer={"branching": True}
    )
    if beta_choice == 0:
        # Pooled model
        beta = pyro.sample("beta", dist.Normal(0, 10))
    elif beta_choice == 1:
        # County specific slopes
        with pyro.plate("num_beta", num_counties):
            beta = pyro.sample("beta", dist.Normal(0, 10))

        beta = beta[..., county]  # Shape: (num_counties,) -> (num_data,)
    elif beta_choice == 2:
        # Partially pooled model
        mean_b = pyro.sample("mean_b", dist.Normal(0, 1))
        std_b = pyro.sample("std_b", dist.Exponential(1))
        with pyro.plate("num_beta", num_counties):
            z_b = pyro.sample("z_b", dist.Normal(0, 1))
        beta = mean_b + std_b * z_b
        beta = beta[..., county]  # Shape: (num_counties,) -> (num_data,)

    theta = alpha + beta * floor_ind
    sigma = pyro.sample("sigma", dist.Exponential(5))
    with pyro.plate("data", log_radon.shape[0]):
        pyro.sample("ys", dist.Normal(theta, sigma), obs=log_radon)
    \end{lstlisting}
    \caption{Pyro program for the radon model.}
    \label{fig:radon_program}
\end{figure}

\subsection{Stacked RJMCMC}

\begin{figure*}[!h]
    \centering
    \includegraphics[width=0.7\textwidth,keepaspectratio]{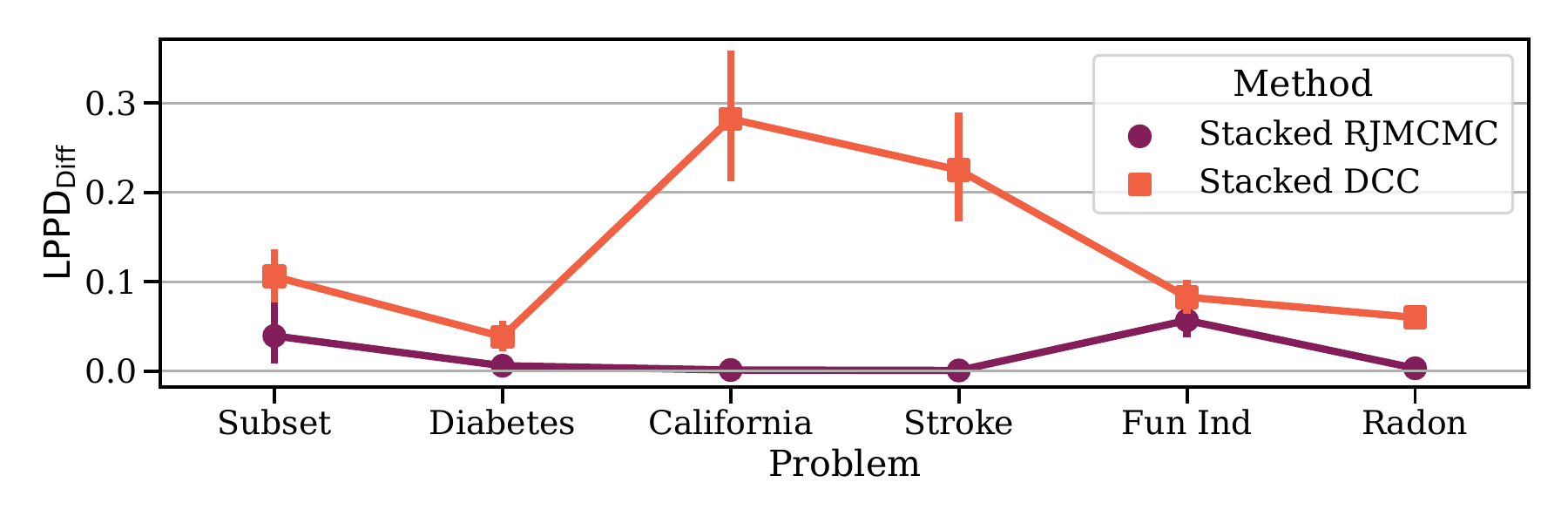}
    \captionof{figure}{Difference in LPPD between RJMCMC and other methods (higher is better).}
    \label{fig:stacked_rjmcmc_lppd}
    \vspace{-10pt}
\end{figure*}

We conducted further experiments to determine the impact of running stacking on top of RJMCMC, we call this \emph{Stacked RJMCMC}.
The results can be viewed in Fig.~\ref{fig:stacked_rjmcmc_lppd} where we plot the difference in LPPD between RJMCMC and other methods, i.e. $\lppd_{\text{Diff}} = \lppd_{\text{Other}} - \lppd_{\text{RJMCMC}}$.
Note, compared to the main paper we here evaluate the difference in LPPD to RJMCMC and not to stacking on top of DCC. 
This is because we care about investigating the performance improvement relative to RJMCMC.
We observe in Fig.~\ref{fig:stacked_rjmcmc_lppd} that Stacked RJMCMC generally leads to higher LPPD than RJMCMC.
The difference is positive for all problem settings. Further, the difference is statisticaly significant under a Wilcoxon-signed rank test in all problems, except for \emph{California} and \emph{Stroke}.

\section{PAC-Bayes and Stacking}
\label{app:pac_bayes}

For completeness, we first give a brief introduction into PAC-Bayes which is mostly based on \citetsupp{morningstar2022PAC}.
\citetsupp{morningstar2022PAC} assume a setting in which data is sampled i.i.d. from $\tilde{y}_{\ell} \sim \ptruedata(\tilde{y})$, that we have a parameterized probability model $p(\tilde{y} \mid \theta)$ and we want to find a mechanism to fit a distribution, $q(\theta)$, over the parameters of the probability model.

The \textbf{true predictive risk} is given by
\begin{equation}
    \mathcal{P}(q) := - \mathbb{E}_{\ptruedata(\tilde{y})}[\log \mathbb{E}_{q(\theta)} [p(\tilde{y} \mid \theta)] ]
\end{equation}
and in most applications is the quantity we care about in the end because it directly measures the quality of our predictions.
However, in practice we cannot evaluate the true predictive risk because we do not have access to the true data generating distribution.
The goal of PAC-Bayes methods is then to provide a stochastic upper bound on $\mathcal{P}(q)$ which can be used to train $q(\theta)$ \citep{masegosa2020learning,morningstar2022PAC}.

The \textbf{empirical predictive risk} 
\begin{equation}
    \overline{\mathcal{P}}(q) := - \frac{1}{L} \sum_{\ell=1}^L \log \mathbb{E}_{q(\theta)} [p(\tilde{y}_{\ell} \mid \theta].
\end{equation}
is an empirical estimate of the true predictive risk.
\emph{Ensemble methods (which include stacking) directly minimize the empirical predictive risk.} 
For example, our stacking objective in Eq.~\eqref{eq:full_stacking_objective_estimated} is a particular instantiation of the empirical predictive risk.
We will explain the connection in more detail below.

The \textbf{true inferential risk}
\begin{equation}
    \mathcal{R}(q) := - \mathbb{E}_{\ptruedata(\tilde{y})}[ \mathbb{E}_{q(\theta)} [\log p(\tilde{y} \mid \theta)] ]
\end{equation}
is an \emph{upper bound on the true predictive risk} (by applying Jensen's inequality).
In the case that our model is well specified, i.e. $\exists\theta'$ s.t. $\ptruedata(\cdot) = p(\cdot \mid \theta')$, then $\argmin \mathcal{P}(q) = \argmin \mathcal{R}(q)$.
Hence, when our model is well specified minimizing the true inferential risk is equivalent to minimizing the true predictive risk.

The \textbf{empirical inferential risk}
\begin{equation}
    \overline{\mathcal{R}}(q) := - \frac{1}{L} \mathbb{E}_{q(\theta)} [\log p(\tilde{y}_{\ell} \mid \theta)] 
\end{equation}
is the empirical estimate of the true inferential risk. 
\emph{Minimizing this risk directly is equivalent to maximum likelihood estimation.}
By adding an extra regularization term to the empirical inferential risk we get the \textbf{PAC-inferential risk}, given by
\begin{equation}
    \widetilde{\mathcal{R}}(q; r, \beta) := \mathbb{E}_{q(\theta)} \left[ - \frac{1}{L} \sum_{\ell=1}^{L} \log p(\tilde{y}_{\ell} \mid \theta) + \frac{1}{\beta L} \log \frac{q(\theta)}{r(\theta)} \right]
\end{equation}
where $r(\theta)$ is a user specified prior distribution on the parameters $\theta$.
This is a stochastic upper bound on the true predictive risk.

\citetsupp{morningstar2022PAC} use results from \citetsupp{burda2015importance} to tighten the PAC-inferential risk and introduce \textbf{the PAC$^M$ bound}
\begin{equation}
    \label{eq:pacm_def}
    \widetilde{\mathcal{P}}_{M,L}(q; r, \beta) := - \frac{1}{L} \sum_{\ell=1}^L \mathbb{E}_{q(\theta^M)} \left[ \log \left( \frac{1}{M} \sum_{j=1}^M p(\tilde{y}_{\ell} \mid \theta_j) \right) \right] + \frac{1}{\beta L} \mathrm{KL}(q(\theta) \parallel r(\theta)).
\end{equation}
Notably, as the PAC-inferential risk, this bound contains a regularization term which is meant to prevent overfitting.

\begin{theorem}[\citetsupp{morningstar2022PAC}]
\label{thm:pacm_bound}
For all $q(\theta)$ absolutely continuous with respect to $r(\theta)$, $\tilde{y}_{\ell} \sim \ptruedata(\tilde{y})$ i.i.d., $\beta \in (0, \infty)$, $L, M \in \mathbb{N}$, $p(\tilde{y} \mid \theta) \in (0, \infty)$ for all $\{\theta \in \Theta \mid \ptruedata(\tilde{y}) > 0 \} \times \{ \theta \in \Theta \mid r(\theta) > 0 \}$, and $\xi \in (0, 1)$, then with probability at least $1 - \xi$,
\begin{equation}
    \mathcal{P}(q) \leq \widetilde{\mathcal{P}}_{M,L}(q; r, \beta) + \psi(\ptruedata, \beta, M, L, r, \xi) - \frac{1}{\beta \, M \, L} \log \xi
\end{equation}
and furthermore (unconditionally)
\begin{equation}
    \widetilde{\mathcal{P}}_{M + 1,L}(q; r, \beta) \leq \widetilde{\mathcal{P}}_{M,L}(q; r, \beta)
\end{equation}
where $ \widetilde{\mathcal{P}}_{M,L}(q; r, \beta)$ as in Eq.~\eqref{eq:pacm_def} and:
\begin{align}
    \psi(\ptruedata, \beta, M, L, r, \xi) &:= \frac{1}{\beta \, M \, L} \log \mathbb{E}_{\ptruedata(\tilde{y}^{L})} \mathbb{E}_{r(\theta^M)}\left[\exp(\beta L M \Delta(\tilde{y}^{L}, \theta^M))\right], \\
    \Delta(\tilde{y}^{L}, \theta^M) &:= \frac{1}{L} \sum_{\ell=1}^L \log\left( \frac{1}{M} \sum_{j=1}^{M} p(\tilde{y}_{\ell} \mid \theta_j)\right) - \mathbb{E}_{\ptruedata(\tilde{y})}\left[ \log \left( \frac{1}{M} \sum_{j=1}^{M} p(\tilde{y} \mid \theta_j) \right) \right].
\end{align}
\end{theorem}
For a proof see Appendix C.3 of \citetsupp{morningstar2022PAC}.

\subsection{From PAC-Bayes to Regularized Stacking}

In order to connect the ideas from \citetsupp{morningstar2021automatic} with the stacking objective for probabilistic programs we need to define a specific form for the parameterized probability model and the distribution $q$ which is meant to be optimized.
We choose the latent variable of our probabilitiy model to be the random variable $k$ which indexes into the SLPs.
Our probabilistic model then turns out to be $p(\tilde{y} \mid k) = \rho_k(\tilde{y})$ and our approximate posterior $q(k)$ is a categorical distribution over $\{1, \dots, K\}$ parameterized by the weights $w$, i.e. $q(k) = \text{Categorical}(w_1, \dots, w_K)$.
In this formulation the true predictive risk is given by
\begin{align}
    \mathcal{P}(q) 
    &= - \mathbb{E}_{\ptruedata(\tilde{y})}[\log \mathbb{E}_{q(k)} [\rho_k(\tilde{y})] ] \\
    &= - \mathbb{E}_{\ptruedata(\tilde{y})}\left[\log \left( \sum_{k=1}^K w_k \rho_k(\tilde{y}) \right) \right]
\end{align}
which is equivalent to Eq.~\eqref{eq:stacking_objective_log_score} in the main paper.
The PAC$^M$ bound becomes
\begin{equation}
    \widetilde{\mathcal{P}}_{M,L}(q; r, \beta) = - \frac{1}{L} \sum_{\ell=1}^N \mathbb{E}_{q(k^M)} \left[ \log \left( \frac{1}{M} \sum_{j=1}^M \rho_{k_j}(\tilde{y}_{\ell}) \right) \right] + \frac{1}{\beta L} \mathrm{KL}(q(k) \parallel r(k)).
\end{equation}
Note that $\beta = 1$ can be interpreted as the ``standard Bayesian'' setting as it provides an equal weighting of the prior term $\mathrm{KL}(q(k) \parallel r(k))$ and likelihood term $\sum_{\ell=1}^N \mathbb{E}_{q(k^M)} \left[ \log \left( \frac{1}{M} \sum_{j=1}^M \rho_{k_j}(\tilde{y}_{\ell}) \right) \right]$.
We can rewrite the sum inside the log as a sum over SLPs, as follows
\begin{equation}
    \widetilde{\mathcal{P}}_{M,L}(q; r, \beta) = - \frac{1}{L} \sum_{\ell=1}^L \mathbb{E}_{q(k^M)} \left[ \log \left( \sum_{k=1}^{K}\frac{\lvert I_k \rvert}{M} \rho_{k}(\tilde{y}_\ell) \right) \right] + \frac{1}{\beta L} \mathrm{KL}(q(k) \parallel r(k))
\end{equation}
where $I_k := \{ k_j \mid j = \{1, \dots, M\}, k_j = k \}$ is the set of all parameter samples that are equal to $k$.
Now by the law of large numbers (LLN) as $m \to \infty$ this converges to 
\begin{equation}
    \widetilde{P}_{\infty,L}(q; r, \beta) = \pacobj(w_{1:K}) = - \frac{1}{L} \sum_{\ell=1}^L \log \left(  \sum_{k=1}^K w_k \rho_{k}(\tilde{y}_{\ell}) \right) + \frac{1}{\beta L} \mathrm{KL}(q(k) \parallel r(k)).
\end{equation}
The first term in this objective is now the stacking objective and the second term is a regularization term pushing the distribution over weights to the prior $r(k)$.
In other words, \textbf{this is Eq.~\eqref{eq:full_stacking_objective_estimated} with an added regularization term.}

There is one caveat with pushing $M \to \infty$, as shown by \citet{morningstar2021automatic} the slack term $\psi$ that controls the tightness of the bound grows linearly in $M$.
Hence, the bound becomes vacuous for infinite $M$.
However, there are several reasons why in our setting it is still reasonable to work with the objective $\widetilde{P}_{\infty,L}(q; r, \beta)$.
First of all, in our specific setting the special case $M = 1$ is the degenerate setting in which we collapse onto a single SLP.
This is exactly the behaviour we are trying to avoid in the first place.
Additionally, \citet{morningstar2022PAC} have shown that performance increases with using larger $M$ and show empirically that using $\widetilde{P}_{\infty,L}(q; r, \beta)$ can be beneficial when it is possible to do so.
The only practical consideration they mention for increasing $M$ are issues with gradient variance which are not applicable in our setting.

To choose the prior one could be inclined to use the posterior SLP weights $Z_k / \sum_{k'} Z_{k'}$, as they are our Bayesian beliefs about which SLP has generated the data. 
However, as we have argued in the main text these weights can be very unstable and expensive to estimate.
Hence, a reasonable default choice could be to use the discrete uniform distribution.
This will further discourage stacking from collapsing onto a single SLP.
For the special case of the prior $r(k)$ being chosen to be the uniform distribution the objective further simplifies to
\begin{equation}
    \pacobj(w_{1:K}) = \, \frac{1}{L} \sum_{\ell=1}^L \log \left(  \sum_{k=1}^K w_k \rho_{k}(\tilde{y}_{\ell}) \right) - \frac{1}{\beta L} \, (\mathrm{H}[q(k; w_{1:K})] + \log K)
\end{equation}
where $\mathrm{H}$ denotes the Shannon entropy.

\begin{figure}[ht]
    \centering
    \begin{lstlisting}[numbers=none, basicstyle=\footnotesize\ttfamily]
def distinct(y):
    model1 = pyro.sample("model1", dist.Bernoulli(0.5))
    if model1:
        z = pyro.sample("z1", dist.Normal(0, 1))
        with pyro.plate("data", y.shape[0]):
            pyro.sample("obs", dist.Normal(z, 0.62), obs=y)
    else:
        z = pyro.sample("z2", dist.Normal(0, 1))
        with pyro.plate("data", y.shape[0]):
            pyro.sample("obs", dist.Normal(z, 2.0), obs=y)
        
def overlap(X, y):
    model1 = pyro.sample("model1", dist.Bernoulli(0.5))
    if model1:
        w = pyro.sample(
            "w1", dist.Normal(0, 1).expand([2]).to_event(1),
        )
        mean = w @ X[:, [0, 2]].T
    elif model2:
        w = pyro.sample(
            "w2", dist.Normal(0, 1).expand([2]).to_event(1),
        )
        mean = w @ X[:, [0, 3]].T
    with pyro.plate("data", X.shape[0]):
        pyro.sample("obs", dist.Normal(mean, 1.0), obs=y)

def dominating(X, y):
    model1 = pyro.sample("model1", dist.Bernoulli(0.5))
    if model1:
        w = pyro.sample("w", dist.Normal(0, 1))
        fs = w * X
    else:
        sin_w = pyro.sample("sin_w", dist.Normal(0, 1))
        fs = torch.sin(sin_w * X)
    sigma = pyro.sample("sigma", dist.Gamma(1, 1))
    with pyro.plate("data", X.shape[0]):
        pyro.sample("obs", dist.Normal(fs, sigma), obs=y)
    \end{lstlisting}
    \caption{Pyro program for experiments in Sec.~\ref{sec:misspecification}}
    \label{fig:small_programs}
\end{figure}